\documentclass[]{fairmeta}

\usepackage{wrapfig}
\usepackage{tabularx}
\usepackage{textcomp}
\usepackage{stfloats}
\usepackage{url}
\usepackage{verbatim}
\usepackage{titlesec}
\usepackage{tocloft}
\usepackage{adjustbox}
\usepackage{multirow}
\usepackage{pifont}
\usepackage{tikz}
\usepackage{comment}
\usepackage{amsmath,amssymb} 
\usepackage{colortbl}  
\usepackage{color}
\usepackage{booktabs} 
\usepackage{hyperref}
\usepackage{graphicx}    
\usepackage{subcaption} 
\usepackage{multirow} 
\usepackage{booktabs} 
\usepackage{subcaption} 
\RequirePackage{xspace}
\makeatletter
\DeclareRobustCommand\onedot{\futurelet\@let@token\@onedot}
\def\@onedot{\ifx\@let@token.\else.\null\fi\xspace}
\usepackage[most]{tcolorbox}
\usepackage{xcolor}
\setcitestyle{square,comma,numbers,sort&compress}

\makeatother

\definecolor{adptorange}{RGB}{248, 205, 172}
\definecolor{cmpblue}{RGB}{189, 215, 238}
\definecolor{cmpblue}{RGB}{189, 215, 238}

\definecolor{our_red}{RGB}{232,157,160}
\definecolor{our_blue}{RGB}{136,206,230}
\definecolor{our_orange}{RGB}{246,200,168}
\definecolor{our_green}{RGB}{178,211,164}

\definecolor{attn_code0}{RGB}{247,215,200}
\definecolor{attn_code1}{RGB}{238,169,139}
\definecolor{mlp_code0}{RGB}{204,201,221}
\definecolor{mlp_code1}{RGB}{102,95,153}

\definecolor{token_blue}{RGB}{84, 120, 140}

\usepackage{pifont}       
\usepackage{bbding}       
\usepackage{fontawesome}
\usepackage{xspace}

\usepackage{float}
\usepackage{multicol}
\usepackage{wrapfig}
\usepackage{dsfont}

\newlength\savewidth

\newcolumntype{x}[1]{>{\centering\arraybackslash}p{#1pt}}
\newcolumntype{y}[1]{>{\raggedright\arraybackslash}p{#1pt}}
\newcolumntype{z}[1]{>{\raggedleft\arraybackslash}p{#1pt}}

\renewcommand{\paragraph}[1]{\vspace{1mm}\noindent\textbf{#1}}
\usepackage{colortbl}
\usepackage{xcolor}
\usepackage{wrapfig}

\newcommand{\ours}{PixelRefer}
\newcommand{\ourslite}{PixelRefer-Lite}

\renewcommand{\paragraph}[1]{\vspace{1.25mm}\noindent\textbf{#1}}

\usepackage{algorithm}
\usepackage{algorithmicx}%
\usepackage{listings}

\definecolor{codeblue}{rgb}{0.25, 0.5, 0.5}
\definecolor{codekw}{rgb}{0.35, 0.35, 0.75}
\lstdefinestyle{Pytorch}{
    language = Python,
    backgroundcolor = \color{white},
    basicstyle = \fontsize{9pt}{8pt}\selectfont\ttfamily\bfseries,
    columns = fullflexible,
    aboveskip=1pt,
    belowskip=1pt,
    breaklines = true,
    captionpos = b,
    commentstyle = \color{codeblue},
    keywordstyle = \color{codekw},
}

\definecolor{green}{HTML}{009000}
\definecolor{red}{HTML}{ea4335}

\newcommand{\huggingface}{\raisebox{-1.5pt}{\includegraphics[height=1.05em]{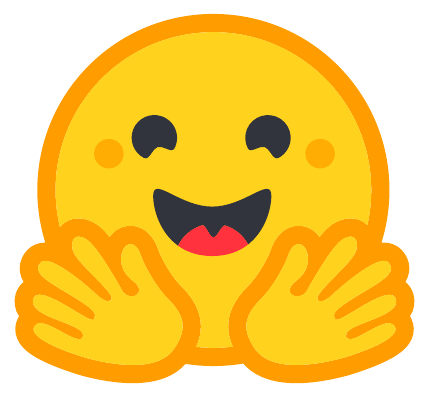}}\xspace}
\newcommand{\github}{\raisebox{-1.5pt}{\includegraphics[height=1.05em]{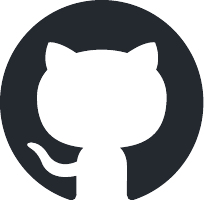}}\xspace}
\newcommand{\worldwideweb}{\raisebox{-1.5pt}{\includegraphics[height=1.05em]{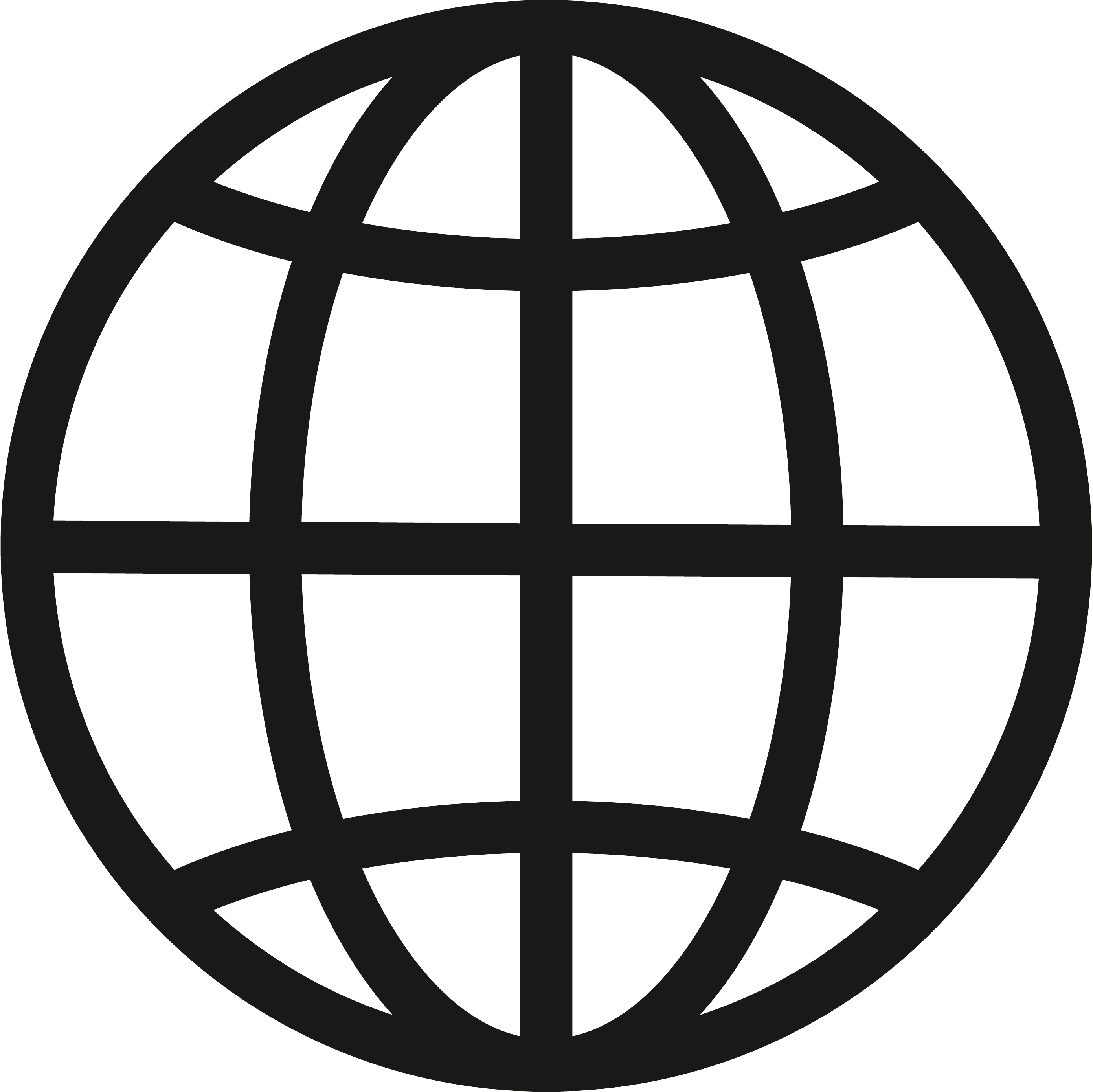}}\xspace}

\title{PixelRefer: A Unified Framework for Spatio-Temporal Object Referring with Arbitrary Granularity}

\author[1]{Yuqian Yuan}
\author[\dagger 1]{Wenqiao Zhang}
\author[2,3]{Xin Li}
\author[4]{Shihao Wang}
\author[2,3]{Kehan Li}
\author[\dagger ]{Wentong Li}
\author[1]{Jun Xiao} 
\author[4]{\\Lei Zhang}
\author[1]{Beng Chin Ooi}

\affiliation[1]{Zhejiang University}
\affiliation[2]{DAMO Academy, Alibaba Group}
\affiliation[3]{Hupan Lab\\}
\affiliation[4]{The Hong Kong Polytechnic University}
\contribution[\dagger]{Corresponding Authors}

\abstract{
Multimodal large language models (MLLMs) have demonstrated strong general-purpose capabilities in open-world visual comprehension.
However, most existing MLLMs primarily focus on holistic, scene-level understanding, often overlooking the need for fine-grained, object-centric reasoning. In this paper, we present \textbf{\ours}, a unified region-level MLLM framework that enables advanced fine-grained understanding over user-specified regions across both images and videos.
Motivated by the observation that LLM attention predominantly focuses on object-level tokens, we propose a Scale-Adaptive Object Tokenizer (SAOT) to generate compact and semantically rich object representations from free-form regions.
Our analysis reveals that global visual tokens contribute mainly in early LLM layers, inspiring the design of \textbf{\ours-Lite}, an efficient variant that employs an Object-Centric Infusion module to pre-fuse global context into object tokens. This yields a lightweight Object-Only Framework that substantially reduces computational cost while maintaining high semantic fidelity.
To facilitate fine-grained instruction tuning, we curate \textbf{PixelRefer-2.2M}, a high-quality object-centric instruction dataset. 
Extensive experiments across a range of benchmarks validate that \ours~achieves leading performance with fewer training samples, while \ours-Lite offers competitive accuracy with notable gains in efficiency. 

\begin{center}
    
    \begin{tabular}{rll}
        \worldwideweb & \textbf{Homepage} & \url{https://circleradon.github.io/PixelRefer} \\
        \huggingface & \textbf{Demo} & \url{https://huggingface.co/spaces/lixin4ever/PixelRefer}\\
        \github & \textbf{Code} & \url{https://github.com/alibaba-damo-academy/PixelRefer}\\
        \huggingface & \textbf{HuggingFace} & \url{https://huggingface.co/collections/Alibaba-DAMO-Academy/pixelrefer} \\
    \end{tabular}
\end{center}
}

\date{\today}

\begin{document}
\thispagestyle{firstheader}
\maketitle
\pagestyle{empty}

\section{Introduction} \label{sec:introduction}
\noindent 
Multi-modal large language models (MLLMs)~\cite{liu2023llava, liu2023improved, gpt4v,bai2023qwen-vl, reid2024gemini1_5, llavanext, lin2024vila, chen2024far} have demonstrated impressive general-purpose capabilities in open-world visual comprehension, spanning both static images and dynamic videos. 
While most existing MLLMs are designed to perform holistic image-level and video-level interpretations and reasoning, they often overlook the need for fine-grained, object-centric understanding, also known as \textit{visual referring}~\cite{kazemzadeh2014referitgame, mao2016generation, yu2016modeling, zellers2019recognition, you2023ferret}, which aims to precisely interpret and reason the semantics of specific, localized regions within visual scenes.
This fine-grained understanding is critical for a wide range of applications that demand accurate object-level comprehension, nuanced event analysis, and reliable predictive reasoning in complex real-world environments, such as human-computer interaction~\cite{mei2025survey}, embodied AI~\cite{yuan2025eoc,dang2025rynnec, cheng2024spatialrgpt}, medical  diagnostics~\cite{alsaad2024multimodal,lin2025healthgpt,zhang2024revisiting,li2025eyecaregpt,xie2025heartcare} and remote sensing  interpretation~\cite{zhang2024earthmarker,shu2025earthmind}.

\begin{figure*}[t]
  \centering
\includegraphics[width=0.999\linewidth]{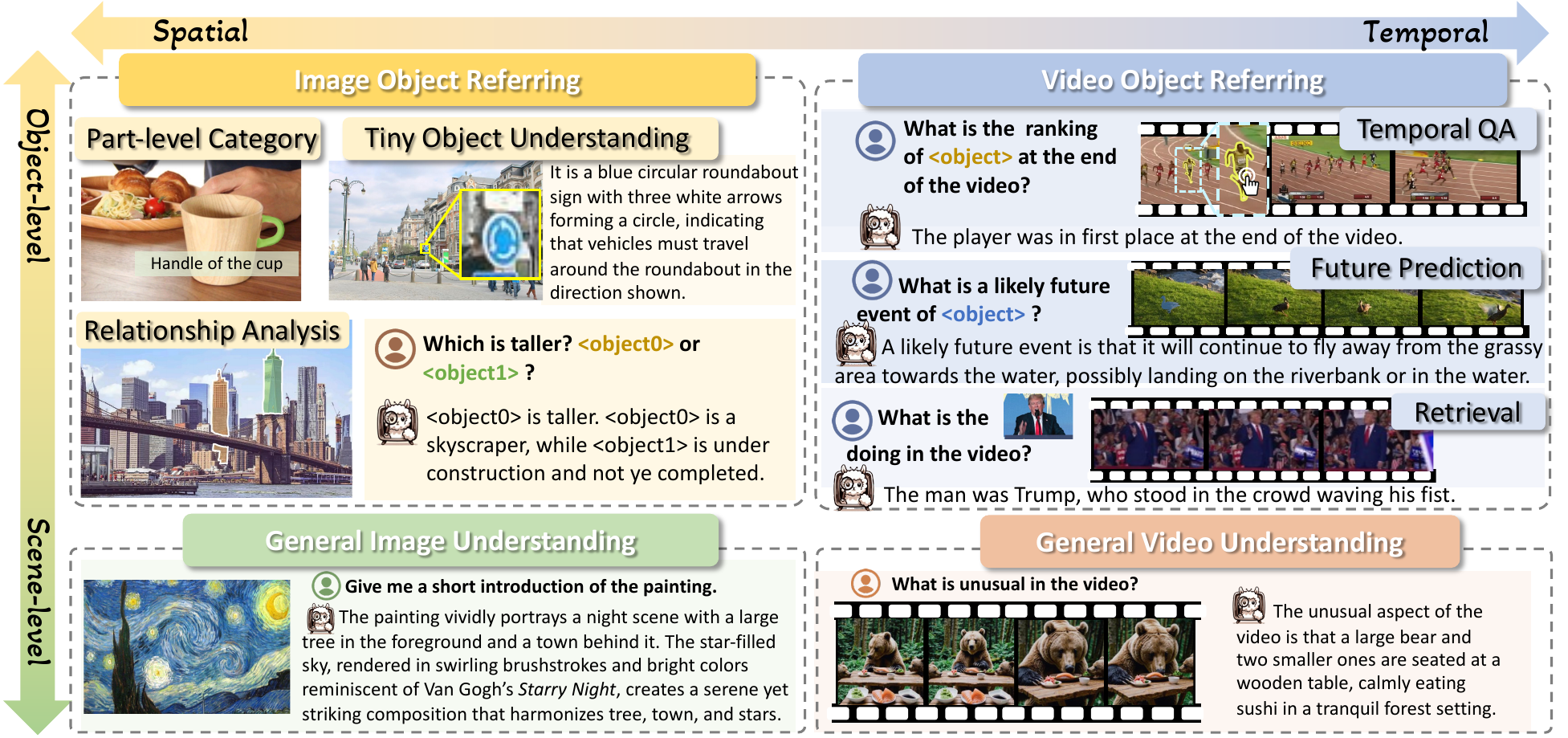}
   \vspace{0.5mm}
   \caption{\ours, a unified region-level MLLM, supports a broad range of tasks at both \textbf{object-level} and \textbf{scene-level}, spanning \textbf{spatial} (images) and \textbf{temporal} (videos) domains. It enables fine-grained spatiotemporal reasoning over user-specified region  with arbitrary semantic granularity, while preserving general-purpose capabilities for holistic visual understanding.}
   \label{fig:intro}
   \vspace{-3.0mm}
\end{figure*}

Early efforts such as SoM~\cite{setofmark} attempt to enhance MLLMs by overlaying visual markers directly onto the input image. 
However, these markers can sometimes be visually ambiguous, blending into the background or obscuring key content, which limits their effectiveness. 
Moreover, modifying object markers requires reprocessing the entire image, 
posing challenges for interaction flexibility and inference efficiency.
To overcome these limitations, a growing number of region-level MLLMs~\cite{zhang2023gpt4roi, yuan2024osprey, guo2024regiongpt, you2023ferret, chen2023shikra, chen2023position, huang2024segment, zhang2024ferret, xuan2024pink, yue2024sc, zhao2023chatspot, rasheed2024glamm, cai2023making, tian2024chatterbox, zhan2024griffon, fei2024vitron, cheng2024spatialrgpt,lin2024draw}  leverage explicit visual prompts or localized queries to extract object-level visual representations, which are aligned with LLMs to enable accurate spatially localized image understanding.
In contrast, the study of  spatiotemporal object understanding in dynamic
videos remains relatively limited. Some works~\cite{wang2024elysium, yu2025merlin} use bounding boxes as text prompts or rely on external RoI tracking~\cite{qiu2024artemis}, but these often yield coarse and temporally inconsistent representations in complex video scenarios.

Recently, research has increasingly shifted toward a unified region-level understanding across both images and videos, aiming to support fine-grained spatiotemporal understanding. 
For instance, the Describe Anything Model (DAM)~\cite{lian2025describe} introduces a focal prompt mechanism to encode user-specified regions and employed a localized vision backbone that integrates global image context into regional representations via gated cross-attention. 
While DAM effectively captures finer details, its architecture is inherently limited to describing a single object at a time, requiring repeated image encoding for multiple regions and thereby incurring substantial computational overhead.
The Perception Anything Model (PAM)~\cite{lin2025perceive} extends SAM 2~\cite{ravisam} by incorporating a learnable semantic perceiver as the interface between vision backbone and LLM. Despite showing promising results, PAM remains largely constrained to captioning tasks,  
limiting its ability to handle more complex reasoning (e.g., object-level QA, multi-object understanding).
Besides, 
its reliance on semantics-agnostic SAM 2 features necessitates large-scale training data (e.g. 8M samples) to achieve sufficient alignment with the LLM.
More critically,
the task-specific architectures of both DAM and PAM undermine the inherent general-purpose capabilities of MLLMs, hindering their flexibility and scalability.

In this work, we revisit the design of a flexible and unified region-level MLLM for fine-grained spatiotemporal object understanding in both images and videos. Unlike prior methods focused primarily on single-object captioning, we advocate  a framework that supports a broad range of object-centric referring tasks, while preserving the general-purpose capabilities of modern MLLMs.
To this end, we argue that such a framework should be built upon a general-purpose MLLM backbone, 
with region-level object representations integrated in a modular and flexible manner,
enabling seamless interaction with the base model without compromising its versatility. 
A preliminary version of this framework was introduced 
in our prior work~\cite{yuan2025videorefer}, where we demonstrated its effectiveness for video-based scenarios. In this paper, we present \textbf{\ours},
a unified region-level MLLM that enables advanced fine-grained understanding over user-specified regions. As illustrated in Fig.~\ref{fig:intro}, {\ours}    supports a variety of perception and reasoning tasks across spatial and temporal dimensions, ranging from object-level to scene-level comprehension. 
Prior to detailing the full model design, we conduct an in-depth preliminary analysis of how our initial framework~\cite{yuan2025videorefer} interprets object-level representations, offering insights that inform the core design of our~\ours.

\begin{figure*}[t]
  \centering
\includegraphics[width=0.99\linewidth]{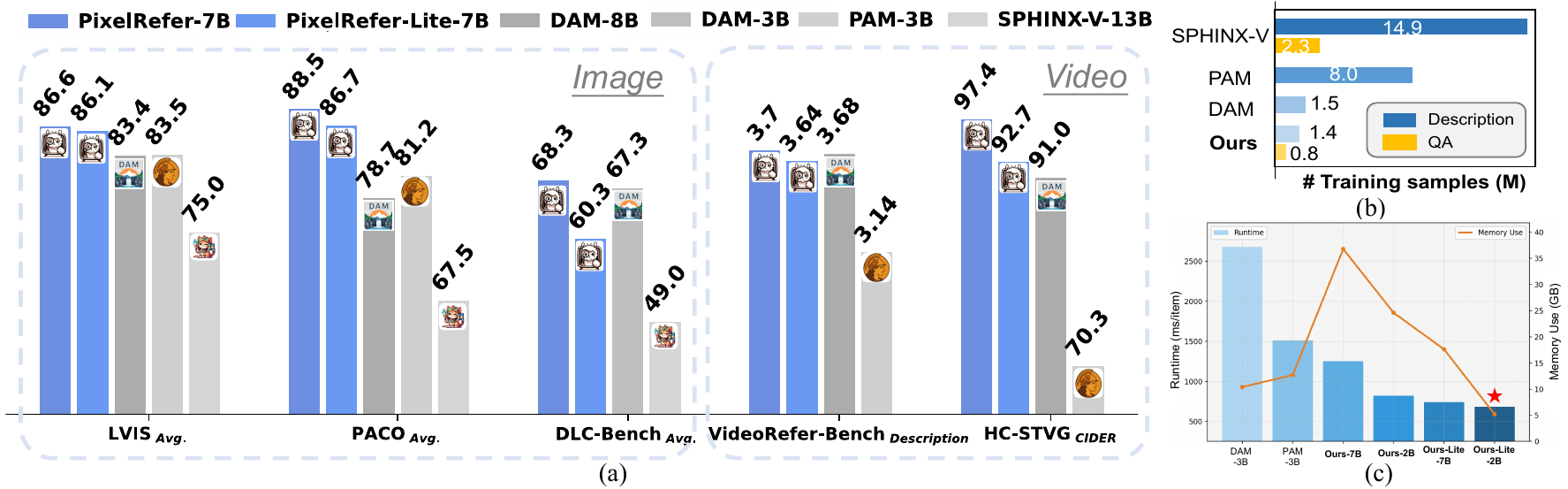}
   \vspace{1.5mm}
   \caption{Quantitative Evaluation and Efficiency Analysis. (a) \textbf{Performance Comparison}: \ours~and \ourslite~ consistently outperform state-of-the-art object-level MLLMs across diverse image (LVIS~\cite{yuan2024osprey}, PACO~\cite{yuan2024osprey}, DLC-Bench~\cite{lian2025describe}) and video  (VideoRefer-Bench, HC-STVG~\cite{tang2021human}) benchmarks.  (b) \textbf{Data Efficiency}: Our method achieves leading performance with fewer training samples compared to existing methods. (c) \textbf{Runtime and Memory Efficiency}: \ourslite~notably reduces inference time and memory usage, clearly demonstrating its efficiency.}
   \label{fig:intro2}
   \vspace{-1.5mm}
\end{figure*}

First, our empirical finding reveals that the LLM's attention is predominantly focused on the region-level tokens corresponding to the referred objects. This highlights the critical role of object tokens quality in determining model performance on object-centric tasks. Motivated by this observation, we introduce the Scale-Adaptive Object Tokenizer (SAOT), a novel object-level tokenizer designed to generate precise and semantically rich region representations. 
It leverages a unified pixel-level mask representation to support arbitrary free-form regions, dynamically adapts to varying object scales, and preserving spatial context, producing compact yet informative object tokens. SAOT is architecture-agnostic and can be seamlessly incorporated into general-purpose MLLM with minimal modifications.

Then, our second empirical finding examines the interaction between global visual tokens and object tokens within the LLM. We observe that attention to global visual tokens is predominantly concentrated in the early layers, while object tokens remain active throughout the LLM. 
However, the global visual tokens contribute excessively to the LLM's overall computational overhead, 
as also noted in prior studies~\cite{fastv,li2024tokenpacker}.
These insights motivate the design of \textbf{\ours-Lite}, an efficient variant of our method based on an \textit{Object-Only Framework}.
Specially, we introduce a lightweight Object-Centric Infusion (OCI) module, which pre-fuses global visual context into object tokens via a hierarchical cross-attention mechanism.  
By retaining only the fused object tokens as input to the LLM, our approach achieves substantial reductions in computational cost while preserving high semantic fidelity.

We further curate a new open-source  dataset, \textbf{{PixelRefer-2.2M}},
structured into two categories: Foundational Object Perception and Visual Instruction Tuning, to support fine-grained alignment between language and both global visual context and local object regions. Extensive experiments are conducted across a wide range of object-centric tasks with varying semantic granularity, including image-level benchmarks such as Category Recognition~\cite{yuan2024osprey}, Phrase-level and Detailed Caption~\cite{ref-l4,vg,lian2025describe}, and Reasoning Questions~\cite{you2023ferret}, as well as video-level benchmarks including VideoRefer-Bench$^\text{D}$~\cite{yuan2025videorefer}, VideoRefer-Bench$^\text{Q}$~\cite{yuan2025videorefer}, and HC-STVG~\cite{tang2021human}. As shown in Fig.~\ref{fig:intro2}-(a)\&(b), our approach consistently achieves state-of-the-art performance, despite being trained on fewer instruction samples than prior advanced counterparts~\cite{lin2024draw, lian2025describe,lin2025perceive}, clearly demonstrating both its effectiveness and data efficiency.
Notably, \ours-Lite delivers competitive accuracy while offering substantial improvements in runtime and memory consumption (Fig.~\ref{fig:intro2}-(c)), highlighting its practicality for real-world applications.

\section{Related Work}  \label{sec: related_work}
\subsection{Multimodal Large Language Models}
Large language models (LLMs) have significantly advanced the field of artificial intelligence by proving their capabilities to tackle diverse tasks related to language comprehension and generation~\cite{minaee2024large}. 
To leverage the potential of LLMs for visual understanding, recent research has focused on 
multimodal LLMs (MLLMs)~\cite{liu2023llava, liu2023improved, gpt4v,bai2023qwen-vl, llavanext,lin2024vila,wang2024qwen2,bai2025qwen2,zhang2025videollama,zhang2024hyperllava}, which integrate vision and language into a unified representation space.
Evolving from image-based MLLMs, recent efforts have explored Video Large Language Models (Video LLMs)~\cite{cheng2024videollama, lin2023video, zhang2023video, maaz2023video, llavanext-video}, aiming to extend multimodal reasoning to dynamic spatiotemporal contexts.
Most Video LLMs empoly pre-trained visual encoders to extract frame-wise or clip-level features, which are then interleaved with textual tokens and processed by LLMs to generate responses~\cite{tang2023video}.
While these models have shown promising progress, they fall short in supporting 
fine-grained spatial and temporal reasoning, especially in object-centric tasks.

\subsection{Region-level Multimodal Large Language Models}
Unlike traditional MLLMs that emphasize  holistic understanding, region-level MLLMs aim for fine-grained, object-centric reasoning. 
Early method like SoM~\cite{setofmark} enhances MLLMs by overlaying visual markers onto the image to indicate object locations, but suffers from ambiguity and limited flexibility.
Recent region-level MLLMs~\cite{zhang2023gpt4roi, yuan2024osprey, guo2024regiongpt, you2023ferret, chen2023shikra, chen2023position, huang2024segment, zhang2024ferret, xuan2024pink, yue2024sc, zhao2023chatspot, rasheed2024glamm, cai2023making, tian2024chatterbox, zhan2024griffon, fei2024vitron} 
introduce explicit visual prompts or region-based queries to generate instance-level representations, improving localized region understanding.
For videos, several  works~\cite{yu2025merlin,wang2024elysium,qiu2024artemis} adopt sparse temporal sampling and coarse object-level references, 
lacking support for multi-object interactions and temporal coherence. 
To address this,  VideoRefer~\cite{yuan2025videorefer}, introduces a simple yet effective architecture for  fine-grained region-level video understanding, supported by large-scale video instruction data and comprehensive benchmarks.
More recently, DAM~\cite{lian2025describe} employs a focal prompt mechanism and localized vision backbone to enable image and video captioning. 
PAM~\cite{lin2025perceive} extends SAM 2~\cite{ravisam} by introducing a semantic perceiver that bridges the visual backbone and LLM, leveraging intermediate SAM 2 features for enhanced region-level understanding.
Despite their promising results, these models remains largely constrained to captioning tasks and fall short in more complex reasoning scenarios.
Moreover, these task-specific architectures compromise the general-purpose nature of MLLMs.

\begin{figure*}[t]
  \centering
\includegraphics[width=0.999\linewidth]{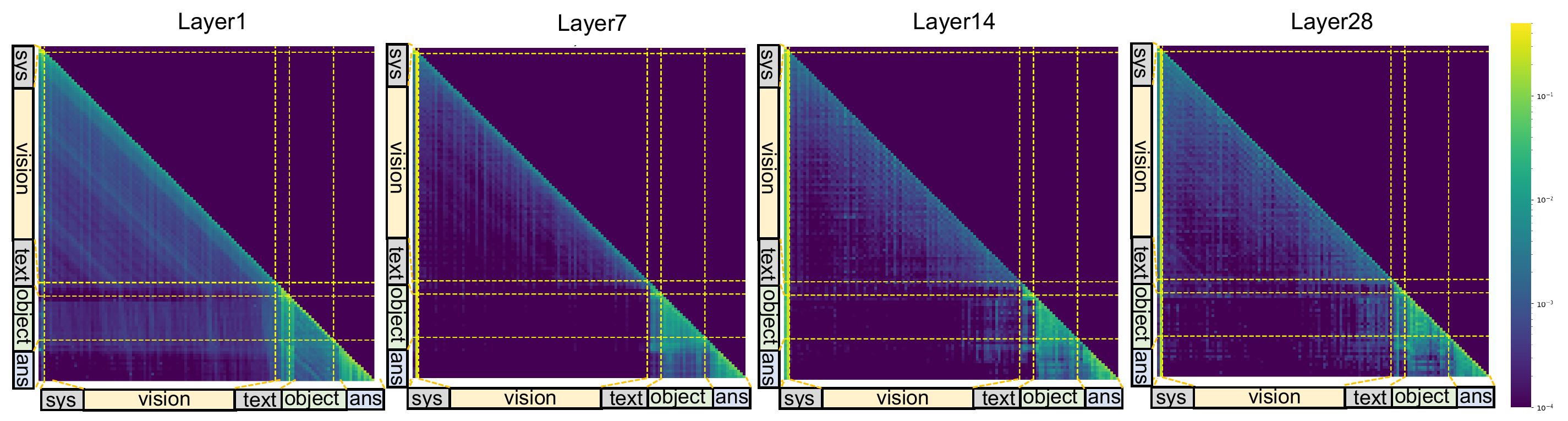}
\vspace{-2.0mm}
   \caption{Visualization of attention maps across different layers (Layer 1, 7, 14 and 28) of the LLM. The input sequence includes system tokens (\texttt{\textbf{sys}}),  global image token (\texttt{\textbf{vision}}), text prompts (\texttt{\textbf{text}}), object-level tokens (\texttt{\textbf{object}}), and answer tokens (\texttt{\textbf{ans}}). For clarity, image tokens are average pooled by a factor of 8. The figure showcases how attention patterns evolve across layers over different tokens.}
   \label{fig:attention}
   \vspace{-1.5mm}
\end{figure*}

\subsection{Benchmarks and Datasets for Region-level MLLMs}
\textbf{Benchmarks.} 
Prior  works~\cite{guo2024regiongpt, huang2024segment, rasheed2024glamm, yu2025merlin,qiu2024artemis} typically assess region-level captioning using traditional language-based metrics~\cite{anderson2016spice, banerjee2005meteor, lin2004rouge, papineni2002bleu, vedantam2015cider}. These metrics 
often measure surface-level textual similarity, fail to reflect factual correctness or fine-grained semantic alignment. 
To address this, recent studies have explored more semantically grounded evaluations. Osprey~\cite{yuan2024osprey} utilizes Sentence-BERT~\cite{reimers2019sentence} to compute sentence-level semantic similarity, along with a semantic IoU metric for word-level alignment. 
Ferret-Bench~\cite{you2023ferret} leverages GPT-4~\cite{openai2023gpt} to score the alignment between predictions and reference captions. 
DLC-Bench~\cite{lian2025describe} further eliminates the reliance on reference captions by scoring model outputs against predefined sets of positive and negative attributes for each region. 
Nevertheless, these benchmarks largely focus on object-level captioning, leaving a notable gap in evaluating spatiotemporal understanding, particularly for complex reasoning in dynamic video scenarios.

\noindent \textbf{Datasets.} 
While several region-level instruction-tuning datasets exist across images~\cite{yuan2024osprey,guo2024regiongpt,lin2024draw,cheng2024spatialrgpt} and videos~\cite{lian2025describe,lin2025perceive}, they predominantly support single-object captions.
This limits their suitability for higher-order visual reasoning tasks like multi-object relationship, and multi-turn QA 
in human-centric, real-world interactions.

\section{How Do MLLMs Understand Object Tokens?}
In this section, we conduct an in-depth investigation into how MLLMs interpret and utilize object-level tokens. Given the complexity of this topic, our preliminary analysis focuses on a  \textit{Vision-Object Framework},  
and examines the role of object tokens within the LLM through attention patterns.

\subsection{Vision-Object Framework}
\label{vision-object}
\begin{figure}[t]
  \centering
\includegraphics[width=0.75\linewidth]{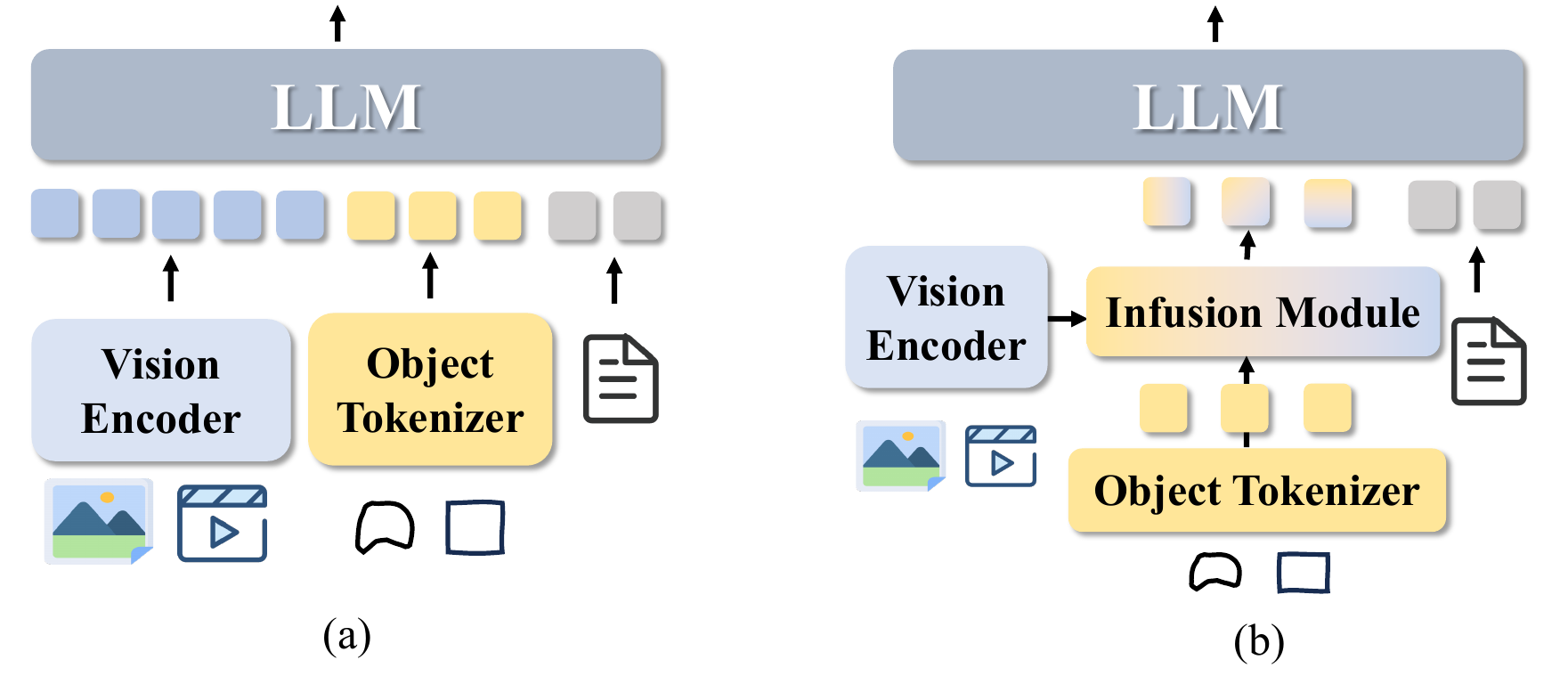}
\vspace{2.0mm}
   \caption{Frameworks of two complementary paradigms 
   for region-level  representations 
   in our approach: (a) illustrates 
   \textbf{Vision-Object Framework}, while (b) presents 
   \textbf{Object-Only Framework}.}
   \label{fig:network}
   \vspace{-2mm}
\end{figure}

As shown in Fig.~\ref{fig:network}-(a), 
the Vision-Object Framework comprises four components: a vision encoder, an object tokenizer, a text tokenizer, and an instruction-following LLM. 

Given a video\footnote{
As images are viewed as single-frame videos, we do not explicitly differentiate between images and videos throughout this work.} $V \in \mathbb{R}^{N \times H \times W \times C}$, where $N$, $H$, $W$, $C$ denote the frame number, height, width and channels, respectively. The vision encoder \( \mathbf{E}_v \) encodes the input and extracts a feature map \( \mathbf{Z} \), which encodes spatial-temporal scene-level information as a sequence of visual tokens \( \mathcal{T}_Z \).  To focus object-centric semantics, we define a set of user-specified region $\boldsymbol{R} = \{R_1, R_2, \ldots, R_n\}$, where $n$ is the number of target objects. Notably, when $n=0$, the framework naturally degenerates to a general visual understanding task.
Each object 
is represented as a collection \( R_j = \{m_{ij}{\mid}i \in \boldsymbol{T} \} \), where \( m_{ij} \in \mathbf{M} \) denotes the binary mask corresponding to a free-form region of interest, and \(\boldsymbol{T}\) being a set containing one or multiple timestamps. The object tokenizer \( \mathbf{E}_R \)  generates enriched object-level representations from $\mathbf{Z}$, yielding object tokens \( \mathcal{T}_R = \mathbf{E}_R({R}, \mathbf{Z}) \).  Finally, the visual tokens \( \mathcal{T}_Z \), object-level tokens \( \mathcal{T}_R \), and linguistic tokens \( \mathcal{T}_X \) are jointly fed into the LLM to generate fine-grained semantic understanding $\mathbf{Y}$. Formally, this process is formulated as:
\begin{equation}
\mathbf{Y} = \Phi(\mathcal{T}_Z, \mathcal{T}_R, \mathcal{T}_X),
\end{equation}where \( \Phi \) denotes the autoregressive decoding function of the LLM. 

The Vision-Object Framework thus enables flexible integration of global scene context, localized object semantics, and linguistic instructions, supporting both fine-grained scene-level and object-level understanding across spatial and temporal dimensions.

\subsection{Preliminary Analyses}
\label{preliminary-analysis}
To gain deeper insights into how object tokens are utilized within the model, we conduct an empirical analysis of attention distributions across LLM layers, spanning from shallow to deep layers.  Representative visualizations are presented in Fig.~\ref{fig:attention}. 
Some key patterns can be observed:

\noindent\textbf{Answer tokens prioritize object tokens.} Across all layers, from shallow to deep layers, answer tokens 
consistently exhibit stronger attention toward object tokens than toward global visual tokens. This pattern indicates that object tokens serve as the primary semantic anchor for question answering. 
Much like human cognition, where specific objects often provide more informative cues than holistic scenes.
Consequently, region-level MLLMs rely on well-structured object representations to generate accurate answers. 
This finding highlights the pivotal role of high-quality object tokenization
in enabling precise and context-aware object-centric understanding.

\begin{wrapfigure}{r}{0.50\textwidth}
  \centering
  \vspace{-2mm}
\includegraphics[width=0.999\linewidth]{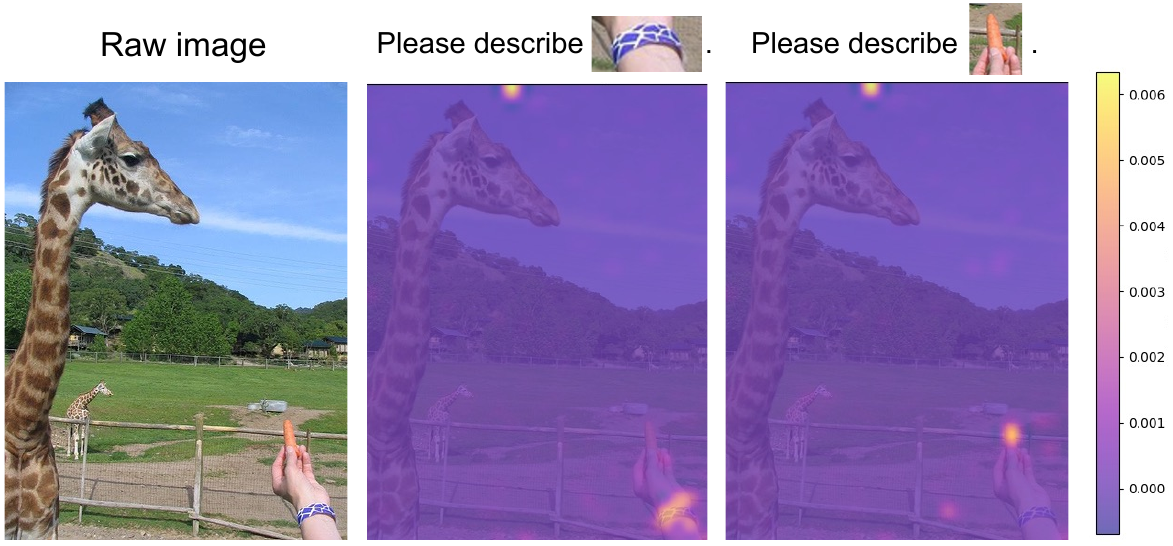}
   \caption{
   Visualization of answer-to-image attention heatmaps for different query regions. The model adaptively focuses on relevant objects while incorporating contextual cues from the surrounding areas.}
   \label{fig:img-attention}
   \vspace{-3.0mm}
\end{wrapfigure}
\noindent\textbf{Answer-to-image token attention is sparse.}
In contrast to object tokens, answer tokens exhibit sparse attention to global image tokens, often manifesting as strip-like patterns in the attention maps.
This suggests that LLMs selectively attend to only a small subset of image tokens deemed relevant to the current query. To assess the interpretability of this selection, 
we further visualize answer-to-image attention distributions across different queried regions
(Fig.~\ref{fig:img-attention}),  
revealing that the model adaptively highlights the corresponding semantically aligned object areas, while occasionally incorporating contextual cues  from broader spatial context (e.g., background or surrounding objects). These findings indicate that global image tokens serve as auxiliary references, complementing the stronger semantic guidance provided by object tokens.

\noindent\textbf{Early fusion of object and image tokens.} As illustrated in Fig.~\ref{fig:attention}, the early layers of the LLM exhibit broad  mutual interaction between object tokens and global image tokens, with attention patterns densely spanning the entire visual token space.
However, as depth increases, the attention gradually shifts, concentrating more heavily on object tokens.
This shift suggests that object tokens serve as compact, information-rich summaries of the relevant visual content, thereby reducing the reliance on raw image tokens in deeper reasoning stages of the LLM. 
This progression from distributed to focused attention reflects a hierarchical processing strategy:
the early layers facilitate comprehensive visual integration by combining both object-specific and global scene features, whereas the later layers  selectively retain task-relevant object-level semantics to support accurate  answer generation.

\section{Methodology} \label{sec:method}
\subsection{Overview}
Motivated by the insights from our attention analysis, we propose two complementary paradigms for object-centric understanding in the our approach. 
The first paradigm, \textbf{PixleRefer}, builds upon the \textit{Vision-Object Framework} (Fig.~\ref{fig:network}-(a)), combining both global visual tokens with object-level tokens. This design enables the model to reason over holistic scene context while leveraging well-constructed object-level representations for more precise and
comprehensive semantic understanding. 
The second paradigm, \textbf{PixelRefer-Lite}, based on the  \textit{Object-Only Framework} (Fig.~\ref{fig:network}-(b)), introduces a  lightweight infusion module that pre-integrates global image context into object tokens prior to LLM processing.
By retaining only object tokens for subsequent LLM decoding, this design
substantially reduces computational cost while preserving strong discriminative capability.

\subsection{PixelRefer}\label{vision-object framework}

As outlined in Sec.~\ref{vision-object}, the Vision-Object Framework consists of four components: a vision encoder, an object tokenizer, a text tokenizer, and an  LLM.
Building on the insights from Sec.~\ref{preliminary-analysis}, our finding highlights the critical role of high-quality object tokenization for enabling precise and context-aware object-centric understanding. To this end, the PixelRefer framework introduces a novel \textit{Scale-Adaptive Object Tokenizer}, 
which employs a dynamic processing strategy to adaptively process objects of varying sizes and shapes. This design ensures robust and 
consistent object-level embeddings across diverse visual inputs.

\subsubsection{Scale-Adaptive Object Tokenizer}
\begin{figure*}[t]
  \centering
\includegraphics[width=0.96\linewidth]{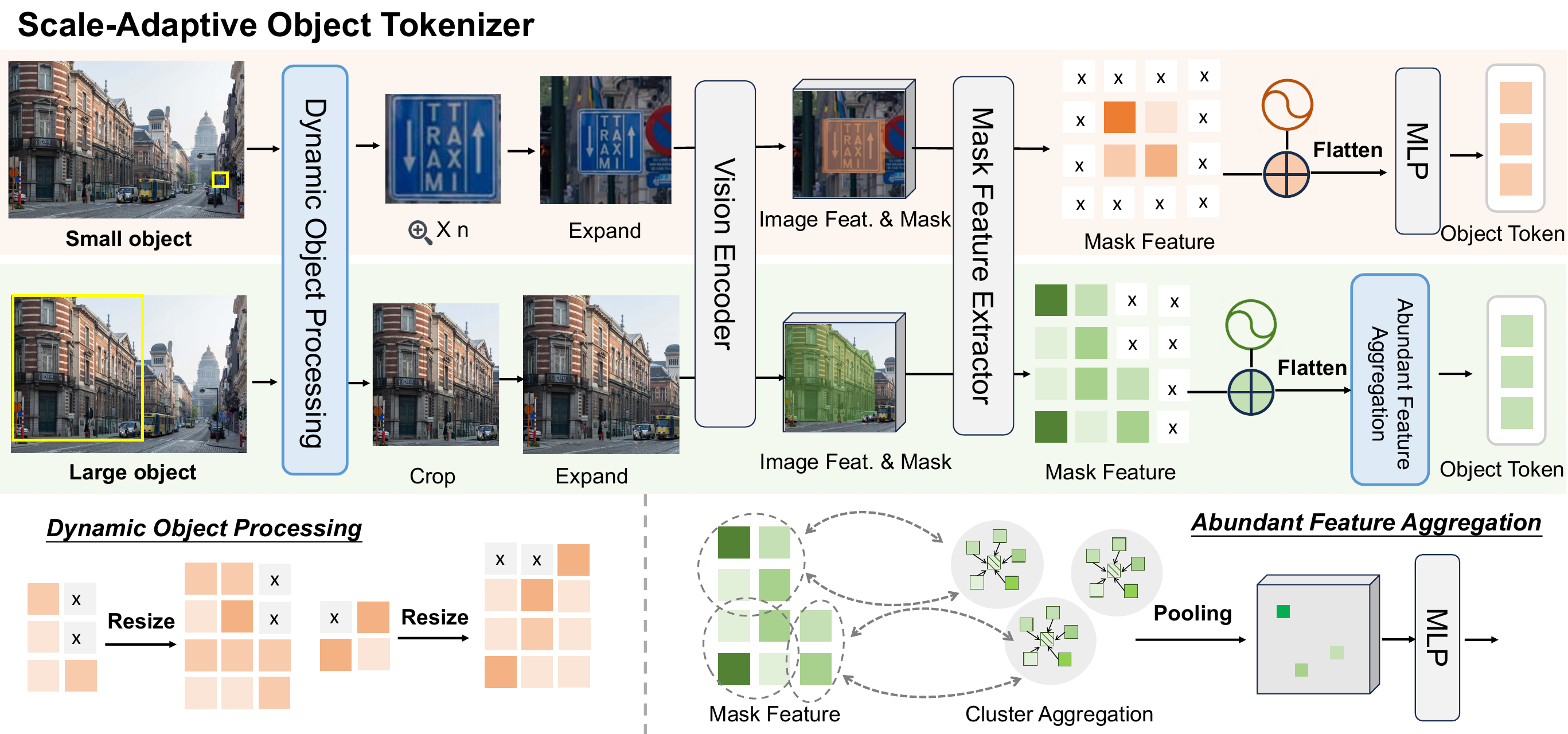}
\vspace{2.5mm}
   \caption{Architecture of our proposed \textbf{Scale-Adaptive Object Tokenizer}. For an input image and a given object, we first perform \textbf{Dynamic Object Processing} to adaptively scale the objects. Subsequently, vision features are extracted from the cropped and expanded sections of the image. To address redundancy prevalent in large objects, we further introduce \textbf{Abundant Feature Aggregation} for efficient feature integration. }
   \label{fig:object_tokenizer}
   \vspace{-1.0mm}
\end{figure*}

In this section, we propose a Scale-Adaptive Object Tokenizer (SAOT) designed to generate accurate and informative object tokens across different scales. Unlike prior approaches that rely on a naive RoI Pooling-based or Mask Pooling-based strategy to encode each region~\cite{zhang2023gpt4roi,yuan2024osprey,yuan2025videorefer,guo2024regiongpt,rasheed2024glamm,fei2024vitron,zhang2025pixel}, our method addresses the common issue of unreliable feature extraction from extremely small or scale-variant regions without comprising fine-grained low-level cues. 
As illustrated in Fig.~\ref{fig:object_tokenizer}, some regions are relatively small, after patchification, may occupy less than a single token, making it difficult to extract reliable features.  
In contrast, our SAOT dynamically adjusts region scale, preserves spatial context, and aggregates redundant features, thereby yielding object tokens that are both accurate and semantically informative.

Given an input image $\mathcal{I} \in \mathbb{R}^{3\times H_I\times W_I}$ and a target region represented by a binary mask $\mathcal{M}$, we perform \textbf{Dynamic Object Processing} to handle scale variations across objects. Specifically, we extract the bounding box $\mathcal{B_R} = (x_b,y_b,w_b,h_b)$ corresponding to the region of interest and adaptively compute a scaling ratio $s$ for region enlargement or reduction:
\begin{equation}
\begin{aligned}
s &= 
\begin{cases} 
\sqrt{\dfrac{\Omega \cdot 100}{|\mathcal{M}|}}, & \text{if } |\mathcal{M}| > 100 \cdot \Omega \\[1.2em]
\sqrt{\dfrac{\Omega \cdot n}{|\mathcal{M}|}}, & \text{elseif } |\mathcal{M}| < n \cdot \Omega \\[1.2em]
1, & \text{otherwise}
\end{cases}
\end{aligned}
\end{equation}where $|\mathcal{M}| = \sum_{i,j} \mathds{1}(\mathcal{M}_{i,j}=1)$ denotes the number of foreground pixels in the mask, and $\Omega = \text{patch}_h \times \text{patch}_w$ is the patch size of the vision encoder. Here, $n$ denotes the number of tokens assigned to each target object.
Once the scaling ratio $s$ is determined, small objects ($|\mathcal{M}| < n \cdot \Omega$) are upscaled by $s$ to retain fine-grained details, wherea large objects ($|\mathcal{M}| > 100 \cdot \Omega$) are downscaled by $s$ to reduce redundancy and computational overhead. 
This scale-adaptive strategy effectively normalizes region sizes across varying object scales, 
ensuring that both small and large objects are encoded with high fidelity while maintaining computational efficiency.

\begin{figure}[t]
  \centering
\includegraphics[width=0.95\linewidth]{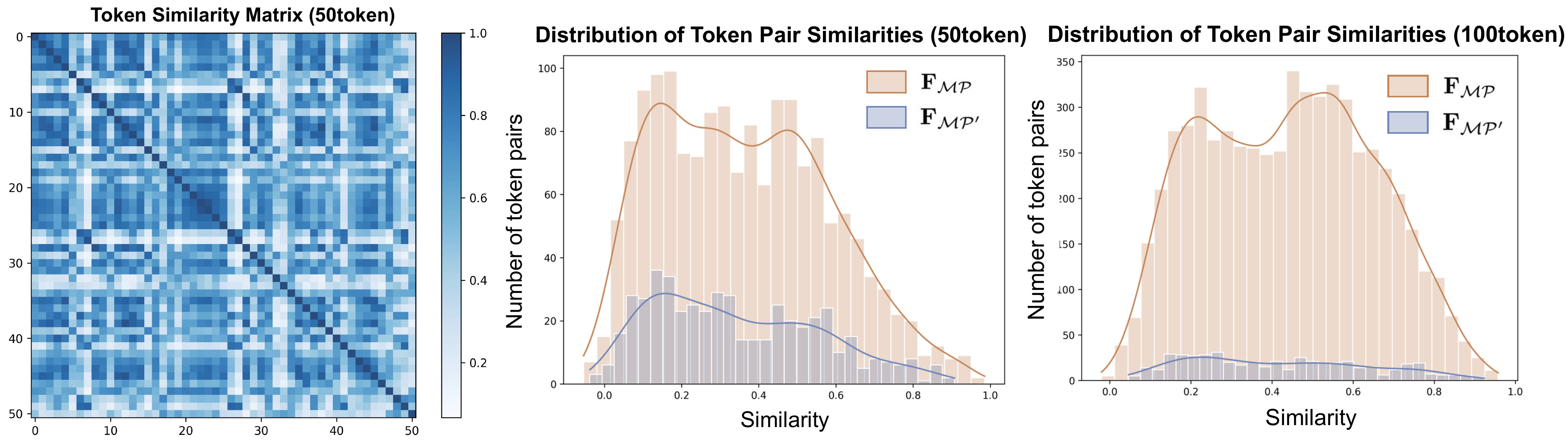}
\vspace{2.5mm}
\caption{Cosine similarity analysis of object tokens. Left: Pairwise similarity matrix of 50 object tokens randomly sampled from a single object.
Right: Histograms of pairwise similarities before ($\mathbf{F}_{\mathcal{MP}}$) and after ($\mathbf{F}_{\mathcal{MP}'}$) \textbf{Abundant Feature Aggregation} for objects with 50 tokens (top) and 100 tokens (bottom), respectively. Aggregated tokens show reduced intra-object similarity, indicating effective redundancy reduction and improved representational compactness.}
\label{fig:token_sim}
\end{figure}

We subsequently apply contextual padding to enlarge the cropped bounding box of the target object,  yielding an expanded region $\mathcal{I_B} \in \mathbb{R}^{3\times h_b'\times w_b'}$. This padded region is then fed into the shared vision encoder to obtain region-level embeddings $\mathbf{F_R}$, which are enriched with contextual information.
To isolate object-specific features, we introduce a \textbf{Mask Feature Extractor}. Specially,  we extract the masked spatial features ${\mathbf{F}_\mathcal{M}}\in \mathbb{R}^{n\times D_I}$ by applying the binary mask $\mathcal{M}$ to the feature map $\mathbf{F_R}$:
\begin{equation}
\mathbf{F}_{\mathcal{M}} = \mathbf{F_R} \odot \mathcal{M}.
\end{equation}Since 
contextual padding disrupts the original spatial alignment of the object within the global image, 
we introduce relative positional encoding to alleviate localization ambiguity:
\begin{equation}
\begin{cases}
p_{i,j}^{(0)} = \big( (j / w_b') \cdot w_b + x_b \big) / (W_I - 1), \\[0.6em]
p_{i,j}^{(1)} = \big( (i / h_b') \cdot h_b + y_b \big) / (H_I - 1),
\end{cases}
\end{equation}
where $0\leq{i}< h_b'$, $0\leq{j}<w_b'$. These coordinates are projected through a linear layer and fused with the masked features to form position-aware object tokens:
\begin{equation}
\mathbf{F}_{\mathcal{MP}} = (\mathbf{F}_{\mathcal{M}} + \text{Linear}\!\left(\mathbf{p}_{i,j}\right))[\mathcal{M}=1].
\end{equation}
As visualized in Fig.~\ref{fig:token_sim}, we observe that the resulting object tokens often exhibit high intra-object similarity, particularly in large or homogeneous regions.
To further mitigate redundancy, we propose an \textbf{Abundant Feature Aggregation}  strategy.
Specifically, we employ k-means clustering to merge redundant tokens: initial $n$ centroids are randomly selected, clustering proceeds for $k$ iterations, and the mean embedding of each cluster $\mathcal{C}_i$ is preserved, resulting in $n$ representative object tokens:
\begin{equation}
\mathbf{F}_{\mathcal{MP}'} = \frac{1}{|\mathcal{C}_i|} \sum_{j\in\mathcal{C}_i} \mathbf{F}_{{\mathcal{MP}}_{j}}, \quad \forall i \in \{1,\dots,n\}.
\label{eq:aggregation}
\end{equation}
Finally, an MLP is applied to generate the final tokens representations of each target object.

\subsection{PixelRefer-Lite}\label{object-only}
As revealed in our preliminary analysis (Sec.~\ref{preliminary-analysis}), 
attention to global visual tokens is primarily concentrated in the early layers of LLM, while object-level tokens maintain strong activation throughout all layers. This observation suggests that semantic fusion between global scene context and object-centric representations is largely completed at shallow layers. However, global visual tokens still comprise the majority of the input sequence of LLM, especially for high-resolution and long video inputs, resulting in significant computational overhead. This inefficiency has also been highlighted in prior studies~\cite{li2024tokenpacker,fastv} as a major bottleneck in MLLMs. 
To improve the overall efficacy of our approach, we introduce \ours-Lite, an efficient variant of our method based on the Object-Only Framework.

\noindent\textbf{Object-Only Framework.} 
The architecture of the Object-Only Framework is illustrated in Fig.~\ref{fig:network}-(b). In contrast to the Vision-Object Framework, which directly concatenates both global visual and object-level tokens as input to the LLM, this framework incorporates a lightweight infusion module to streamline visual processing.
Specially, the infusion module integrates global visual context \( \mathcal{T}_Z \) into object tokens \( \mathcal{T}_R \), 
enabling each object to be enriched with global contextual cues. This design significantly reduces the total number of vision tokens passed to the LLM while preserving critical semantic content.
Formally, the infusion module is defined as:
\begin{equation}
\mathcal{T}_O = \Psi(\mathcal{T}_R, \mathcal{T}_Z),
\end{equation}where \( \Psi \) denotes the infusion function.  
Here, \( \mathcal{T}_O \) represents the enhanced object tokens that are integrated with scene-level visual context.
Subsequently, the refined object tokens \( \mathcal{T}_O \) are concatenated with linguistic tokens \( \mathcal{T}_X \) and fed into the LLM for decoding, yielding precise context-aware object-level semantic understanding:
\begin{equation}
\mathbf{Y} = \Phi(\mathcal{T}_O, \mathcal{T}_X),
\end{equation}where $\Phi$ denotes the LLM's decoding function. By eliminating the need to retain dense global visual tokens, the Object-Only Framework offers a token-efficient yet semantically rich alternative, particularly well-suited for processing high-resolution images or long video sequences.

\noindent\textbf{Object-Centric Infusion Module.} Within our PixelRefer-Lite framework, we introduce an Object-Centric Infusion (OCI) module designed to 
hierarchically integrates contextual visual information into object tokens, thereby enhancing their semantic representations. 
To model broader contextual understanding based on  long-range dependencies, 
OCI module adopts a two-step cross-attention infusion strategy that progressively incorporates local and global visual context into the object tokens.
In the first step, \emph{Local-to-Object Attention}, fine-grained visual embeddings extracted from  locally expanded image regions are injected into the object tokens. 
This operation enables refined tokens to capture detailed local context from the object's immediate  surroundings, preserving fidelity to object's original appearance while becoming more robust against occlusion or noise.
In the second step, \emph{Global-to-Object Attention}, object tokens are further conditioned on scene-level embeddings derived from the raw image.
This global integration introduces long-range dependencies and holistic semantics, complementing the previously injected local details and enabling more scene-aware object understanding.
The detailed processing flow of the proposed OCI module is presented in Algorithm~\ref{alg1}. This hierarchical injection mechanism mirrors human perception, where object recognition is progressively refined by situating local detail within its broader scene context.
For implementation, we adopt standard attention operations for cross-attention, 
enabling direct compatibility with recent advances in efficient attention kernels~\cite{dao2023flashattention}.

\begin{algorithm}[t]
\caption{\small Object-Centric Infusion Module}
\footnotesize
\textbf{Input:} Object tokens $\mathcal{T}_R \in \mathbb{R}^{n \times D}$; 
Raw image $\mathcal{I} \in \mathbb{R}^{3\times H_I\times W_I}$; Object mask $m\in \mathbb{R}^{ H_I\times W_I}$\\
\textbf{Output:} Fused object tokens $\mathcal{T}_O \in \mathbb{R}^{n \times D}$
\begin{algorithmic}[1]
    \State $F_l \gets \mathbf{E}_v\big(\text{Resize}(\text{LocalCrop}(\mathcal{I}, m))\big)$ \Comment{\textcolor{gray}{Extract local features}}
    \State $\hat{\mathcal{T}}_R \gets \mathrm{LN}(\mathcal{T}_R)$
   \State $\mathcal{T}_l \gets \mathcal{T}_R + \text{Local-to-Object Attn}(\hat{\mathcal{T}}_R, F_l, F_l)$ 
   \State \Comment{\textcolor{gray}{Pre-Norm + Residual}}

    \State $F_g \gets \mathbf{E}_v\big(\text{Resize}(\mathcal{I})\big)$ \Comment{\textcolor{gray}{Extract global features}}
    \State $\hat{\mathcal{T}}_l \gets \mathrm{LN}(\mathcal{T}_l)$
   \State $\mathcal{T}_O \gets \mathcal{T}_l + \text{Global-to-Object Attn}(\hat{\mathcal{T}}_l, F_g, F_g)$ 
   \State \Comment{\textcolor{gray}{Pre-Norm + Residual}}
\end{algorithmic} 
\textbf{Return} $\mathcal{T}_O$
\label{alg1}
\end{algorithm}

\noindent\textbf{Extension to Videos. } Given that a video can be regarded as a sequence of images across different timestamps, the Object-Only Framework naturally generalizes to the video domain by processing temporally ordered sequences of object tokens, each associated with its respective frame.
To incorporate temporal information, we prepend timestamps embeddings to each object token, enabling the model to distinguish objects across different frames.
Each object mask is independently processed through the proposed OCI module. For object mask extraction, we employ SAM 2~\cite{ravisam} to generate high-quality segmentation masks on sampled video frames.

\section{Dataset}
With the growing demand in fine-grained, pixel-level object understanding, recent studies have developed instruction-tuning datasets to advance this task, including image-level data~\cite{yuan2024osprey,guo2024regiongpt,lin2024draw} and video-level data~\cite{lian2025describe,lin2025perceive}. However, most existing datasets remain limited to single-object semantic captioning for object-level recognition, especially in video scenes, which falls short in supporting complex visual reasoning required for human-centric video interactions in real-world scenarios.
To address this gap, we introduce VideoRefer-700K in our prior work~\cite{yuan2025videorefer}, a meticulously curated large-scale region-text video instruction dataset. It features 
region-level descriptions and multi-turn question–answer (QA) pairs spanning basic inquiries, compositional reasoning, and future event prediction. 
Beyond this, we further carefully collect diverse open-source image-level and video-level datasets and systematically organize them into two categories: Foundational Object Perception and Visual Instruction Tuning, thereby providing a robust knowledge foundation for model fine-tuning.

\subsection{Data Collection}
We curate a collection of open-source datasets and systematically organize them into two categories: Foundational Object Perception and Visual Instruction Tuning.

\subsubsection{Foundational Object Perception Data}
\begin{figure}[t]
  \centering
\includegraphics[width=0.999\linewidth]{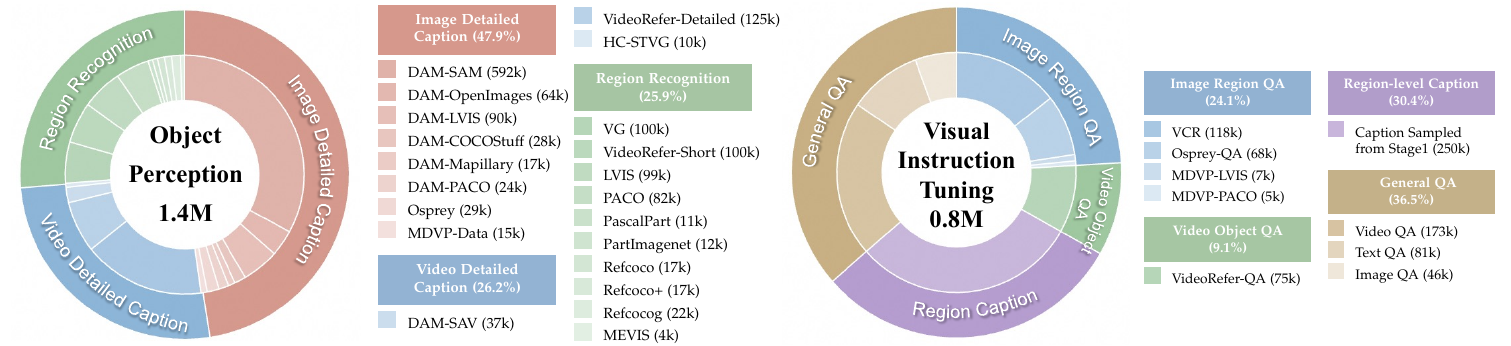}
\vspace{1.0mm}
   \caption{Overview of datasets used for model training. Left: Data distribution for Foundational Object Perception training (1.4M samples). Right: Data used for Visual Instruction Tuning (0.8M samples).
   }
   \label{fig:data_overview}
   \vspace{-2.5mm}
\end{figure}

While pre-trained LLMs and Vision Transformers (ViTs) encode rich general priors about the world, they lack precision at the regional level. 
To address this limitation, we first strengthen fine-grained regional alignment through carefully curated supervision before advancing to instruction tuning. 
Figure~\ref{fig:data_overview} illustrates the composition of data in this stage, totaling 1.4M samples across three complementary categories.

\noindent\textbf{Region Recognition.} 
A critical foundation for high-quality visual understanding lies in region recognition. To this end, we curate a multi-scale cognitive dataset encompassing objects, parts, and temporal dynamics. Object-level annotations from LVIS~\cite{gupta2019lvis}, Visual Genome~\cite{vg}, and RefCOCO/RefCOCO+~\cite{yu2016modeling} are strategically combined with fine-grained part-level datasets (PACO~\cite{ramanathan2023paco}, Pascal-Part~\cite{chen2014detect}, PartImageNet~\cite{he2022partimagenet}), enabling hierarchical learning from whole objects to constituent parts. Furthermore, temporal alignment is introduced through VideoRefer-Short captions and MEVIS~\cite{ding2023mevis}, bridging static region understanding with dynamic scene perception.

\noindent\textbf{Regional Image Detailed Caption.} For region-level image captioning, we integrate diverse domain data sources to enhance descriptive richness. Specifically, we aggregate DAM~\cite{lian2025describe} samples constructed from SAM, OpenImages, LVIS, COCOStuff, Mapillary, and PACO, ensuring broad coverage of object categories and contexts. To further expand descriptive variety, we incorporate Osprey-caption and MDVP-Data, which contribute more fine-grained narrative supervision.

\noindent\textbf{Regional Video Detailed Caption.}
Compared with image data, detailed video captions remain scarce yet essential for modeling temporal and contextual understanding. To fill this gap, we leverage three complementary resources: our self-constructed VideoRefer-Detailed captions, DAM-SAV~\cite{lian2025describe} dataset, and HC-STVG~\cite{tang2021human}.

\subsubsection{Visual Instruction Tuning Data}
Visual instruction tuning aims to endow LMMs with the ability to understand, follow, and respond to natural-language instructions grounded in visual inputs. Achieving this requires supervision that is not only large in scale but also rich in instruction diversity and reasoning depth. To this end, we curate four complementary types of data, as illustrated in Figure~\ref{fig:data_overview}, totaling 0.8M samples.

\noindent\textbf{Image Region QA.} To strengthen localized reasoning capabilities at the region level, we incorporate Osprey-QA~\cite{yuan2024osprey}, MDVP-LVIS~\cite{lin2024draw}, and MDVP-PACO~\cite{lin2024draw}, all annotated using GPT-4/GPT-4o to ensure high-quality, instruction-rich supervision. In addition, we include VCR~\cite{zellers2019recognition} data, which extends beyond recognition toward commonsense reasoning over visual contexts, pushing the model to infer intent, causality, and social cues from images.

\noindent\textbf{Video Object QA.}
Compared with images, object-centric QA in videos remains limited. 
To address this gap, we constructed 75K VideoRefer-QA samples using a multi-agent pipeline~\cite{yuan2025videorefer}, enabling precise question–answer supervision over dynamic objects in temporal sequences. This enriches the model’s capacity to reason about object identity, attributes, and interactions across time, an essential skill for video understanding.

\noindent\textbf{Region Caption.}
To consolidate the model’s ability for regional captioning, we sample 250K captions from the data in Foundational Object Perception. This prompts continuity between foundational perception pretraining and instruction tuning, aligning descriptive generation with QA-based reasoning. By jointly incorporating caption-style and QA-style supervision, we balance generative expressiveness and discriminative reasoning.

\noindent\textbf{General QA.}
To broaden the model's instruction-following capabilities, we introduce general visual QA beyond region-specific tasks.
These are sampled from sources such as LLaVA-Video~\cite{zhang2025llava} and LLaVA-OV~\cite{li2024llava}, providing a wide range of open-ended queries. This component equips the model with the flexibility to handle heterogeneous instruction types, ranging from factual recognition to complex visual reasoning.

\section{Experiment} \label{sec: experiment}

\subsection{Experiment Setup}

\noindent\textbf{Implementation Details.}
Our base model is built upon VideoLLaMA 3~\cite{zhang2025videollama},
a robust unified architecture capable of understanding both images and videos. 
Its vision encoder processes images with dynamic resolutions using Rotary Position Embedding (RoPE)~\cite{su2024roformer}, enabling fine-grained perception of small regions compared to strategies based on fixed input sizes. 
For video inputs, the model efficiently reduces the number of vision tokens by leveraging their similarity, resulting in more precise and compact video representations. 
We initialize our model with the VideoLLaMA 3-Image weights.
Benefiting from large volume of high-quality vision-text pre-training data, VideoLLaMA 3 exhibits strong and resilient vision understanding capabilities.
Additionally, we adopt a progressive 
training strategy consisting of two stages: foundational object perception training (Stage~1) and visual instruction tuning (Stage~2). The global batch size is configured to 256, and each stage is trained for one epoch. A cosine learning rate scheduler is employed across all stages, with a warm-up ratio of 0.03 applied to the learning rate. In Stage~1, the learning rates are configured as follows: $1 \times 10^{-5}$ for the LLM and projector, and $1 \times 10^{-3}$ for the object encoder. In Stage~2, the learning rate for all parameters is uniformly set to $1 \times 10^{-5}$.
The number of object tokens $n$ is set to 32 in the main experiments.
For efficiency analysis, experiments are carried out using one NVIDIA A100 80GB GPU.

\noindent\textbf{Efficiency Metrics.} 
To comprehensively assess model efficiency, we adopt three key metrics:
FLOPs, GPU memory usage and inference time, providing a holistic view of computational complexity.
For the calculation of  FLOPs, as in FastV~\cite{fastv},  we calculate the FLOPs for the multi-head attention and the feed-forward network (FFN) modules  as $4nd^2+2n^2d+2ndm$.  Here, $n$ denotes the token count, $d$ is the hidden state size, and $m$ represents FFN's intermediate dimension.  Considering the projection in \texttt{k} and \texttt{v} is not equal to \texttt{q}, and there are three linear layer in FFN of Qwen-style LLM, the FLOPs is modified as $2nd^2+2ndd_{kv}+2n^2d+3ndm$.
For our Vision-Object Framework, the FLOPs are calculated by:
{\small
\begin{equation}
\begin{split}
    \text{FLOPs} 
    &= \sum^{S} K_s ( 2(L_R+L_Z)d^2 + 2(L_R+L_Z)d d_{kv} \\
    &\quad\quad\quad+ 2(L_R+L_Z)^2d + 3(L_R+L_Z)dm).
\end{split}
\end{equation}
}For our Object-Only Framework, FLOPs are computed as:
{\small
\begin{equation}
\begin{split}
    \text{FLOPs} 
    &=\sum^{S} K_s \left( 2L_Rd^2 + 2L_Rd d_{kv} + 2L_R^2d + 3L_Rdm \right) \\
    &\quad + 2(L_R+L_{Z_L})d^2 + 2(L_R+L_{Z_G})d^2, 
\end{split}
\end{equation}
}where $L_R$, $L_Z$, $L_{Z_L}$ and $L_{Z_G}$ denote the numbers of region tokens, vision tokens, local-to-object tokens, and global-to-object tokens, respectively.

\begin{table*}[t]
\caption{\textbf{Performance on image-level region understanding benchmarks:} including \textbf{category-level} (LVIS and PACO), \textbf{detailed captioning} (DLC-Bench and Ref-L4 [CLAIR]), \textbf{phrase-level} (Ref-L4 and VG) and \textbf{reasoning-level} (Ferret-Reasoning). The best results are \textbf{bold} and the second-best results are \underline{underlined}.}
\vspace{1.5mm}
\centering
\resizebox{0.99\textwidth}{!}{
\begin{tabular}{l|cc|cc|ccc|ccc|cc|c}
\toprule 
          \multirow{2}{*}{\textbf{Method}}& \multicolumn{2}{c|}{\textbf{LVIS}} & \multicolumn{2}{c|}{\textbf{PACO}} & \multicolumn{3}{c|}{\textbf{DLC-Bench}} & \multicolumn{3}{c|}{\textbf{Ref-L4}} & \multicolumn{2}{c}{\textbf{VG}}  & \textbf{Ferret}\\ 
          \cmidrule(lr){2-3} \cmidrule(lr){4-5} \cmidrule(lr){6-8} \cmidrule(lr){9-11} \cmidrule(lr){12-13} \cmidrule(lr){14-14}
          & \textbf{SSim}   & \textbf{SIoU}  & \textbf{SSim}   & \textbf{SIoU}  & \textbf{Pos}  & \textbf{Neg} & \textbf{Avg}  & \textbf{CLAIR} & \textbf{METEOR} & \textbf{CIDER} & \textbf{METEOR} & \textbf{CIDER} & \textbf{Reasoning} \\ \midrule
DAM-3B~\cite{lian2025describe}    & -- &  --  &  -- &  -- & \textbf{52.3} & 82.2 & \underline{67.3} &     --  & \underline{17.2}  & 56.4 & -- & -- & -- \\
PAM-3B~\cite{lin2025perceive}    & 88.6  & 78.3  & 87.4  & 74.9  &  --  & -- & -- & -- & \underline{17.2} & 59.7 & \textbf{20.8} & 142.3 & --\\
Osprey-7B~\cite{yuan2024osprey} & 65.2 & 38.2 & 73.1 & 52.7 &  --  & -- & -- & -- & -- & -- & -- & -- & 67.8 \\
Ferret-7B~\cite{you2023ferret} & 63.8 & 36.6 & 58.7 & 26.0 & 14.2 & 46.8 & 30.5 & 45.2 & 10.7 & 39.7 & -- & -- & 67.3 \\
DAM-8B~\cite{lian2025describe}    & 89.0  & 77.7  & 84.2  & 73.2  & --  & -- &  --  & 57.9  & \textbf{19.4}   & 70.0 & -- & -- & --\\
SPHINX-V-13B~\cite{lin2024draw} & 87.1 & 62.9 & 79.9 & 55.0 & 26.3 & 71.6 & 49.0 & 51.2 & 10.7 & 32.4 & \underline{20.6} & 141.8 & 70.4 \\
\rowcolor{blue!5}\textbf{\ourslite-2B} & 89.4  & 82.0  & 89.3 & 81.9 & 41.1 & 80.2 & 60.7 & \textbf{63.5} & 13.0 & 92.9 & 18.6 & 155.0 & 74.5\\
\rowcolor{blue!5}\textbf{\ourslite-7B} & 89.6 & \underline{82.5} & \underline{90.3} & \underline{83.1} & 48.2 & 72.4 & 60.3 & 56.2 & 12.6 & 89.4 & 19.0 & \underline{161.4} & 78.1 \\
\rowcolor{blue!5}\textbf{\ours-2B} & 89.8  & \underline{82.5}  & \underline{90.1}  & \underline{82.7}  & 46.8 & \underline{85.4} & 66.1 & \underline{60.9}& 14.1 & \textbf{102.2} & 19.7 & 161.2 & \underline{78.5} \\
\rowcolor{blue!5}\textbf{\ours-7B} & \textbf{90.5}  & \textbf{82.7}  & \textbf{91.7}  & \textbf{85.3}  & \underline{49.6} & \textbf{87.0} & \textbf{68.3} & 60.8 & 13.8 & \underline{98.2} & 19.7   & \textbf{168.2} & \textbf{83.1} \\
\bottomrule 
\end{tabular}}
\label{tab:image-bench}
\end{table*}

\subsection{Main Results}

\subsubsection{Image-level Benchmarks}

\begin{figure}[t]
  \centering
\includegraphics[width=0.999\linewidth]{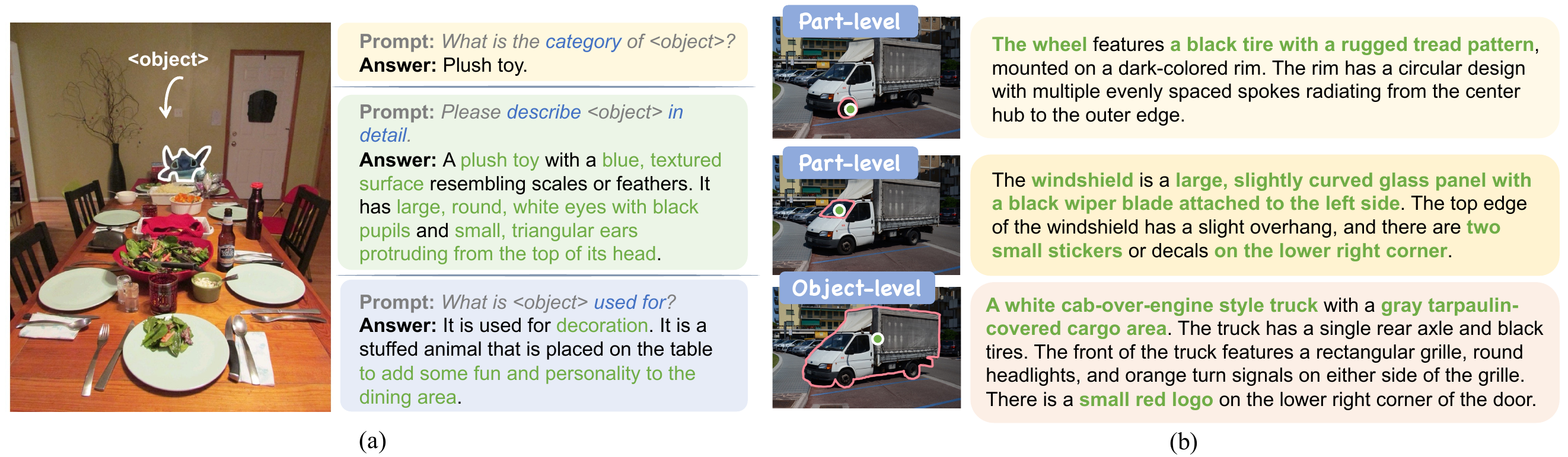}
\vspace{0.1mm}
   \caption{(a) \textbf{Multi-perspective object understanding with \ours.} The model generates diverse responses according to different prompts. (b) \textbf{Granular visual understanding with \ours. }
   \ours~yields distinct detailed descriptions at part-level and object-level based on specified region granularity.
   }
   \label{fig:vis_image}
   \vspace{-4.0mm}
\end{figure}

We begin by evaluating the model on image-level region understanding benchmarks, which encompass three key aspects: category recognition, phrase-level captioning, and detailed captioning. Table~\ref{tab:image-bench} summarizes the comparison results. 
Additionally, we provide visualization examples in Fig.~\ref{fig:vis_image}-(a) to demonstrate the model’s adaptability to instructions, which showcases varied responses based on different prompts. In Fig.~\ref{fig:vis_image}-(b), we highlight the model’s capability to offer diverse, detailed descriptions of each region, with different granularity from object-level to part-level details.

\noindent\textbf{Category Recognition.} This task requires the model to output the category or part-level category corresponding to a given region. 
Following Osprey~\cite{yuan2024osprey}, we adopt object-level LVIS~\cite{gupta2019lvis} and part-level PACO~\cite{ramanathan2023paco} as  evaluation benchmarks. Our approach achieves state-of-the-art (SOTA) performance on both datasets. 
On the PACO benchmark, a particularly challenging category recognition task  involving both whole objects and object parts in complex scenes, which requires the model to distinguish whether a region corresponds to an object or a part, our \ours-7B attains 91.7\% semantic similarity (SSim) and 85.3\% semantic IoU (SIoU), surpassing the previous best by 4.3\% and 10.4\%, respectively. In addition, our lightweight \ourslite-7B also surpasses the previous SOTA by 2.9\% SSim and 8.4\% SIoU. 
Notably, the PACO dataset is dominated by small part-level regions.
The substantial performance gains in these part regions highlight the effectiveness of our proposed Scale-Adaptive Object Tokenizer, particularly in handling fine-grained, small-scale visual components.

\noindent\textbf{Phrase-level Caption.} This task requires the model to generate a short phrase or brief description for each given region. 
We evaluated performance on VG~\cite{vg} and Ref-L4~\cite{ref-l4} datasets. 
Our \ours-7B model achieves comparable performance to existing methods on VG-METEOR, and attains the best performance on VG-CIDER, with a score of 168.2\%, outperforming PAM-3B~\cite{lin2025perceive} by 45.9\%. 
For Ref-L4, we conduct zero-shot evaluation on the Objects365~\cite{shao2019objects365} split. Following the evaluation protocol of ~\cite{lian2025describe}, both model predictions and ground-truth captions are first summarized by GPT-4o~\cite{gpt4o}, and then evaluated using short captioning metrics.
In this setting, our \ours-2B achieves a 32.2\% improvement in CIDER over the previous best model. While its METEOR score is slightly lower, this may be attributed to formatting mismatches between the generated outputs and  ground-truth annotations.

\noindent\textbf{Detailed Caption.} In this setting, the model is expected to generate  comprehensive and fine-grained descriptions of each region, going beyond short phrases to capture nuanced attributes and contextual information.
To assess this capability, we conduct evaluations on DLC-Bench~\cite{lian2025describe} and Ref-L4 (CLAIR)~\cite{ref-l4} benchmarks on  the   Objects365 subset.
Our models demonstrate strong performance.  
In particular, \ours-7B achieves state-of-the-art results on DLC-Bench with 68.3\%, while \ourslite-2B surpasses the previous best model on Ref-L4-CLAIR by 5.6\%.

\noindent\textbf{Reasoning Questions.} In this setting, the model is required to perform reasoning based on one or more referred regions correctly.
We evaluate this ability on the Referring Reasoning task of Ferret-Bench~\cite{you2023ferret}, which involves commonsense reasoning in visual context. Our \ours~demonstrates superior performance on this task, improving the score from 70.4\% to 83.1\% (+12.7\%). These results indicate that our approach effectively narrows the gap between visual perception and high-level reasoning, enabling more accurate interpretation of complex visual scenarios.

\begin{table}[t]
\begin{minipage}{0.49\linewidth}
\caption{Performance comparisons on VideoRefer-Bench$^\text{D}$. The best results are \textbf{bold} and the second-best results are \underline{underlined}. For general baselines,  masks of the targets are overlaid on the original video. S: single-frame mask, M: multi-frame masks.}
\vspace{1.0mm}
\centering
\resizebox{\columnwidth}{!}{
\begin{tabular}{l|c|ccccc}
\toprule
 \textbf{Method} & \textbf{Mode} & \textbf{SC} & \textbf{AD} & \textbf{TD} & \textbf{HD} & \textbf{Avg.} \\ 
\midrule
\rowcolor[HTML]{F2F2F2} \multicolumn{7}{l}{\textbf{Generalist Models}} \\ \midrule
LongVU-7B~\cite{shen2024longvu} & S & 2.02 & 1.45 & 1.98 & 1.12 & 1.64 \\ 
LongVA-7B~\cite{zhang2024longva} & S & 2.63 & 1.59 & 2.12 & 2.10 & 2.11  \\
LLaVA-OV-7B~\cite{li2024llava}& S & 2.62 & 1.58 & 2.19 & 2.07 & 2.12 \\
Qwen2-VL-7B~\cite{wang2024qwen2} & S & 2.97 & 2.24 & 2.03 & 2.31 & 2.39 \\
InternVL2-26B~\cite{chen2024far} & S & 3.55 & 2.99 & 2.57 & 2.25 & 2.84 \\
GPT-4o~\cite{gpt4o} & S & 3.34 & 2.96 & 3.01 & 2.50 & 2.95 \\
GPT-4o-mini~\cite{gpt4o} & S & 3.56 & 2.85 & 2.87 & 2.38 & 2.92 \\
\midrule
\rowcolor[HTML]{F2F2F2} \multicolumn{7}{l}{\textbf{Region-level Models}} \\
 \midrule
DAM-3B~\cite{lian2025describe} & M & 3.62 & 2.86 & 2.81 & 2.67 & 2.99 \\
PAM-3B~\cite{lin2025perceive} & S & 3.92 & 2.84 & 2.88 & 2.94 & 3.14 \\
Elysium-7B~\cite{wang2024elysium} & S & 2.35 & 0.30 & 0.02 & \textbf{3.59} & 1.57 \\
Artemis-7B~\cite{qiu2024artemis} & S & 3.42 & 1.34 & 1.39 & 2.90 & 2.26 \\
VideoRefer-7B~\cite{yuan2025videorefer} & S & 4.44 & 3.27 & 3.10 & 3.04 & 3.46 \\
DAM-8B~\cite{lian2025describe} & M & \underline{4.69} & \textbf{3.61} & \underline{3.34} & 3.09 & \underline{3.68} \\
\rowcolor{blue!5}\textbf{\ourslite-2B} & M & 4.56 & 3.41 & 3.08 & 3.12 & 3.53 \\
\rowcolor{blue!5}\textbf{\ourslite-7B} & 
M & 4.69 & 3.56 & 2.28 & 3.06 & 3.64 \\
\rowcolor{blue!5}\textbf{\ours-2B} & S & \underline{4.59} & 3.40 & 3.25 & 3.09 & 3.58 \\
\rowcolor{blue!5}\textbf{\ours-7B} & S & \textbf{4.70} & \underline{3.59} & \textbf{3.39} & \underline{3.13} & \textbf{3.70} \\
\bottomrule
\end{tabular}}
\vspace{-2mm}
\label{tab:videorefer-d}
\end{minipage}
\hfill
\begin{minipage}{0.49\linewidth}
\caption{Performance comparisons on VideoRefer-Bench$^\text{Q}$.  BQ: Basic Questions, SQ: Sequential Questions, RIQ: Relationship Questions,RsQ: Reasoning Questions, FP: Future Prediction.}
\vspace{1.0mm}
  \centering
  \resizebox{\columnwidth}{!}{
  \begin{tabular}{l|c|c|c|c|c|c}
    \toprule
    \textbf{Method} & BQ & SQ & RlQ & RsQ & FP & Avg. \\
    \midrule
    \rowcolor[HTML]{F2F2F2} \multicolumn{7}{l}{\textbf{Generalist Models}} \\ 
    \midrule
    LongVU-7B~\cite{shen2024longvu} & 47.2 & 61.3 & 57.5 & 85.3 & 65.8 & 61.0 \\
    LongVA-7B~\cite{zhang2024longva} & 56.2 & 62.5 & 52.0 & 83.9 & 65.8 & 61.8 \\
    InternVL2-26B~\cite{chen2024far} & 58.5 & 63.5 & 53.4 & 88.0 & 78.9 & 65.0 \\
    GPT-4o-mini~\cite{gpt4o} & 57.6 & 67.1 & 56.5 & 85.9 & 75.4 & 65.8 \\
    Qwen2-VL-7B~\cite{yang2024qwen2} & 62.0 & 69.6 & 54.9 & 87.3 & 74.6 & 66.0 \\
    LLaVA-OV-7B~\cite{li2024llava} & 58.7 & 62.9 & 64.7 & 87.4 & 76.3 & 67.4\\
    GPT-4o~\cite{gpt4o} & 62.3 & \underline{74.5} & 66.0 & 88.0 & 73.7 & 71.3 \\
    \midrule
    \rowcolor[HTML]{F2F2F2} \multicolumn{7}{l}{\textbf{Region-level Models}} \\ 
    \midrule
    Osprey-7B~\cite{yuan2024osprey} & 45.9 & 47.1 & 30.0 & 48.6 & 23.7 & 39.9 \\
    Ferret-7B~\cite{you2023ferret} & 35.2 & 44.7 & 41.9 & 70.4 & 74.6 & 48.8 \\ 
    Elysium-7B~\cite{wang2024elysium} & - & - & - & - & - & - \\
    Artemis-7B~\cite{qiu2024artemis} & - & - & - & - & - & - \\
    PAM-3B~\cite{lin2025perceive}  & - & - & - & - & - & - \\
    DAM-8B~\cite{lian2025describe} & - & - & - & - & - & - \\
    VideoRefer-7B~\cite{yuan2025videorefer}  & 75.4 & 68.6 & 59.3 & 89.4 & 78.1 & 71.9 \\
    \rowcolor{blue!5} \textbf{\ourslite-2B} & 70.3 & 58.8 & 56.4 & 80.7 & 73.7 & 65.7 \\
    \rowcolor{blue!5} \textbf{\ourslite-7B} & 81.2 & 72.7 & \underline{68.5} & 88.4 & \textbf{81.6} & \underline{76.9} \\
    \rowcolor{blue!5} \textbf{\ours-2B} & \underline{82.1} & 73.0 & 64.7 & \textbf{90.2} & \textbf{81.6} & 76.5 \\
    \rowcolor{blue!5} \textbf{\ours-7B} & \textbf{84.5} & \textbf{76.9} & \textbf{71.5} & \underline{89.5} & \underline{79.7} & \textbf{79.4} \\
    \bottomrule
  \end{tabular}}

  \label{tab:videorefer-bench-q}
  \end{minipage}

\end{table}

\subsubsection{Video-level Benchmarks}
\begin{figure}[t]
  \centering
\includegraphics[width=0.999\linewidth]{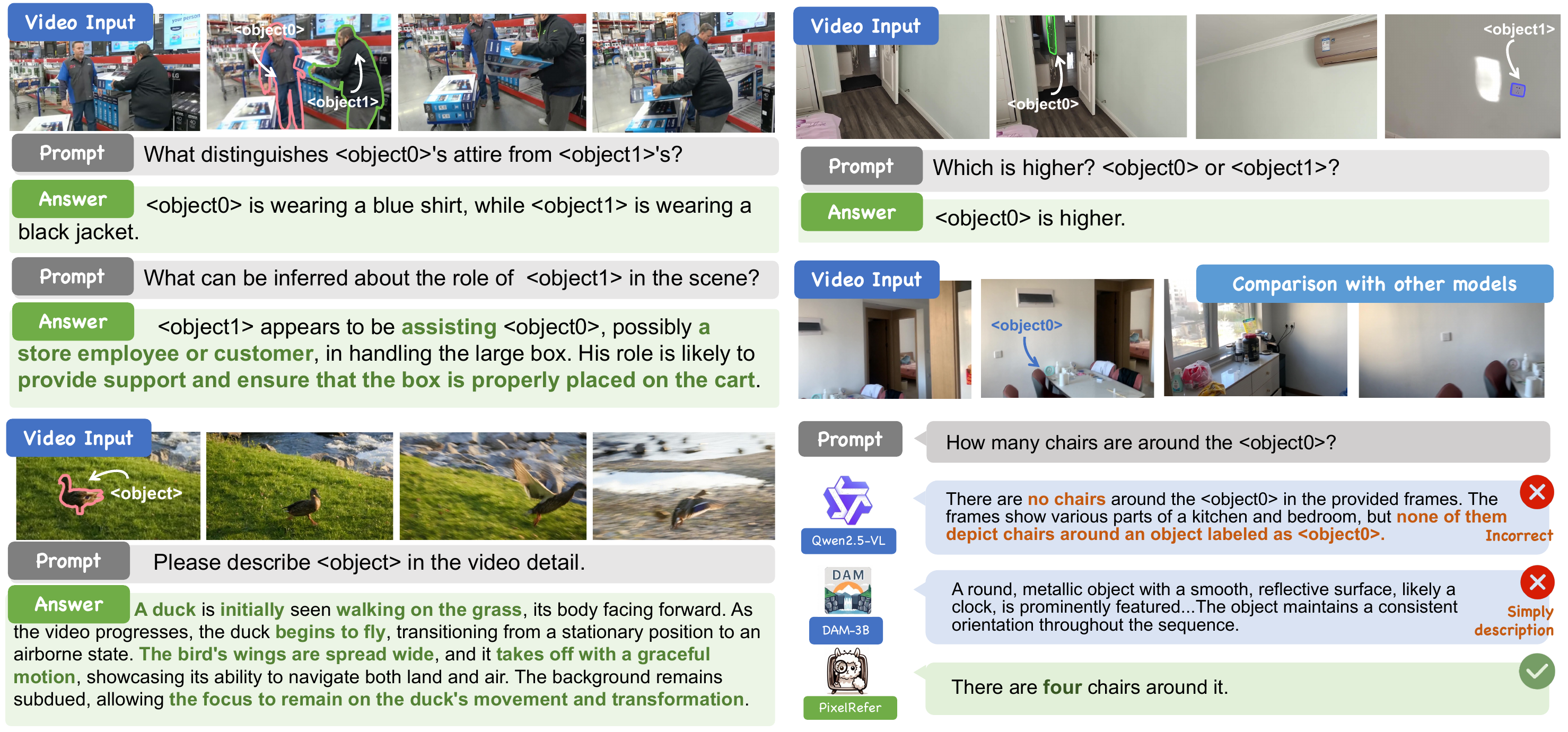}
\vspace{1.0mm}
   \caption{Left: \textbf{Versatile video referring with \ours.}~\ours~handles diverse video referring tasks, including video object captioning, multi-object question answering, and zero-shot spatial understanding. Right: \textbf{Comparing ~\ours~with Qwen2.5-VL~\cite{bai2025qwen2}, DAM~\cite{lian2025describe} on video object referring task.} ~\ours~exhibits the ability to accurately identify specific objects while also comprehending the overall context of the video.}
   \label{fig:vis_video}
   \vspace{-2.0mm}
\end{figure}

To thoroughly evaluate video-level object understanding, we conduct experiments on both caption-level and question-answering (QA)-level subtasks, leveraging both existing established benchmarks and ours newly constructed VideoRefer-Bench designed for this study.
For caption-level tasks, we employ VideoRefer-Bench$^\text{D}$ and HC-STVG~\cite{tang2021human}. For QA-based tasks that 
require answering dynamic and context-aware queries, we adopt challenging
VideoRefer-Bench$^\text{Q}$.
This benchmark enables us to evaluate 
how effectively models can infer answers to complex questions, capturing spatiotemporal dependencies and contextual understanding in video-based scenarios.
Qualitative results are presented in Fig.~\ref{fig:vis_video}. Our method,~\ours, exhibits strong capability across diverse video referring tasks, including video object captioning, multi-object question answering, and zero-shot spatial understanding.

\begin{table}[t]
\begin{minipage}{0.5\linewidth}
\caption{Quantitative comparisons with video object-centric methods on HC-STVG benchmark.}
\vspace{1.0mm}
\centering
\resizebox{\columnwidth}{!}{
\begin{tabular}{l|ccccc}
\toprule
 \textbf{Method} & \textbf{METEOR} & \textbf{CIDER} & \textbf{BLEU@4} & \textbf{ROUGE-L} & \textbf{SPICE} \\
\midrule
DAM-3B~\cite{lian2025describe} & 18.2 & 72.7 & -- & -- & -- \\
PAM-3B~\cite{lin2025perceive} & \textbf{23.3} & 70.3 & -- & -- & -- \\
Elysium-7B~\cite{wang2024elysium} & -- & -- & -- & -- & -- \\
Merlin-7B~\cite{yu2025merlin} & 11.3 & 10.5 & 3.3 & 26.0 & 20.1 \\
Artemis-7B~\cite{qiu2024artemis} & 18.0 & 53.2 & 15.5 & 40.8 & 25.4 \\
VideoRefer-7B~\cite{yuan2025videorefer} & 18.7 & 68.6 & -- & -- & -- \\
DAM-8B~\cite{lian2025describe} & 21.0 & 91.0 & 19.8 & 45.9 & \underline{31.4} \\
\rowcolor{blue!5}\textbf{\ourslite-2B} & 21.1 & 91.3 & 19.0 & 45.8 & 31.2 \\
\rowcolor{blue!5}\textbf{\ourslite-7B} & \underline{21.9} & \underline{92.7} & \underline{20.7} & \textbf{46.5} & 31.3 \\ 
\rowcolor{blue!5}\textbf{\ours-2B} & 19.5 & 78.9 & 17.2 & 43.8 & 30.1 \\
\rowcolor{blue!5}\textbf{\ours-7B} & 21.1 & \textbf{97.4} & \textbf{20.1} & \underline{46.1} & \textbf{32.5} \\
\bottomrule
\end{tabular}}
\vspace{-2.0mm}
\label{tab:hc-stvg}
\end{minipage}
\hfill
\begin{minipage}{0.48\linewidth}
\centering
\caption{FLOPs and memory consumption of different VideoRefer model variants under image and video settings. Experiments are conducted on DLC-Bench~\cite{lian2025describe} (Image) and HC-STVG~\cite{tang2021human} (Video).}
\label{tab:flops}
\resizebox{\columnwidth}{!}{%
\begin{tabular}{l|cccc|cc}
\toprule
\textbf{Method} & 
$L_R$ & $L_Z$  & $L_{Z_G}$ & $L_{Z_L}$ &\textbf{FLOPs(T)$\downarrow$} & \textbf{Memory(GB)$\downarrow$} \\
\midrule
\rowcolor[HTML]{F2F2F2} \multicolumn{7}{l}{\textbf{DLC-Bench} (\textbf{Image})} \\ 
\midrule
\ours-2B      & 32 & $\sim 1408$ & -- & -- & 1.51 & 13.2 \\
\rowcolor{blue!5} \ours-2B-Lite & 32 &  0 & 576 & 256 & 0.03 & 4.9 \\
\ours-7B      & 32 & $\sim 1408$ & -- & --  & 7.08 & 25.1 \\
\rowcolor{blue!5} \ours-7B-Lite & 32 & 0 & 576 & 256 & 0.17 & 15.8 \\
\midrule
\rowcolor[HTML]{F2F2F2} \multicolumn{7}{l}
{\textbf{HC-STVG} (\textbf{Video})} \\ 
\midrule
\ours-2B & 32 & $\sim 7185$  & -- & -- & 11.15  & 24.6  \\
\rowcolor{blue!5} \ours-2B-Lite & 32 & 0 & 576 & 256 & 0.11 & 5.1\\
\ours-7B  & 32 & $\sim 7185$  & -- & -- & 43.83 &  36.9 \\
\rowcolor{blue!5} \ours-7B-Lite & 32 & 0 & 576 & 256 & 0.61 & 17.6 \\
\bottomrule
\end{tabular}}
\end{minipage}
\end{table}

\noindent\textbf{VideoRefer-Bench$^\text{D}$.}
We benchmark our method on VideoRefer-Bench$^\text{D}$ and compare it against several advanced generalist models, including GPT-4o~\cite{gpt4o}, GPT-4o-mini~\cite{gpt4o}, InternVL2~\cite{chen2024far}, Qwen2-VL~\cite{yang2024qwen2}, LLaVA-OV~\cite{li2024llava}, LongVA~\cite{zhang2024longva}, and LongVU~\cite{shen2024longvu}, as well as region-level specialist models for object-level understanding, such as Elysium~\cite{wang2024elysium}, Artemis~\cite{qiu2024artemis}, DAM~\cite{lian2025describe} and PAM~\cite{lin2025perceive}. 
In the single-frame (S) mode, we use the first frame containing the target object and its aligned boundary as input for generalist models, while image-level methods process a random frame paired with the corresponding region prompt. In the multi-frame (M) mode, object masks are generated for each key frame using the off-the-shelf SAM 2~\cite{ravi2024sam}.
For our \ours, we simply sample a random single frame in the single-frame (S) mode. Table~\ref{tab:videorefer-d} presents the comparison results. Our approach achieves leading average performance in regional-temporal video understanding. 
Notably, in the single-frame setting, \ours~achieves top scores of 4.70 in Subject Correspondence (SC), 3.59 in Appearance Description (AD), and 3.39 in Temporal Description (TD), surpassing DAM-8B~\cite{lian2025describe} by an average of +0.02, despite the latter leveraging multi-frame inputs with denser object masks.

\noindent\textbf{VideoRefer-Bench$^\text{Q}$.}
We further evaluate our method on VideoRefer-Bench$^\text{Q}$, which assesses a model’s ability to answer multiple-choice questions involved in referred video regions.
Notebaly, some specialist models like DAM~\cite{lian2025describe}, PAM~\cite{lin2025perceive}, Elysium~\cite{wang2024elysium} and Artemis~\cite{qiu2024artemis} lack the capability to support this task.
Therefore, we compare our method against generalist models as well as image-based region-level baselines to provide a comprehensive performance analysis.
As presented in Table~\ref{tab:videorefer-bench-q}, our approach achieves the best average performance, scoring 79.4\%, and exceeding the closed-source GPT-4o by 8.1\%. Additionally, \ours-7B achieves top scores across multiple subcategories, including Basic Questions (BQ) at 84.5\%, Sequential Questions (SQ) at 76.9\%, Relationship Questions (RQ) at 71.5\%, Reasoning Questions (RQ) at 89.5\%, and Future Prediction (FP) at 79.7\%. These results clearly validate  the effectiveness of our method in addressing the challenges of spatiotemporal video understanding.  
The lightweight \ourslite~variant shows relatively lower performance, primarily due to architectural constraints that prevent it from leveraging global scene-level choice-based training data, thereby limiting its instruction-following capability.

\noindent\textbf{HC-STVG~\cite{tang2021human}.}
This benchmark assesses a model's ability to generate detailed object-level descriptions in videos, with a particular focus on human-centric scenarios. As reported in Table~\ref{tab:hc-stvg}, our proposed {\ours }     achieves leading performance, surpassing the previous best model, DAM-8B~\cite{lian2025describe}.
Specifically, \ours-7B attains 21.1 METEOR (+0.1), 97.4 CIDER (+6.1), 20.1 BLEU@4 (+0.3), 46.1 ROUGE-L (+0.2), and 32.5 SPICE (+1.1), compared to DAM-8B~\cite{lian2025describe}, demonstrating consistent improvements across all  metrics. In addition, \ourslite-7B delivers performance comparable to \ours-7B, while offering greater efficiency.


\subsection{Efficiency Analysis}

\begin{wraptable}{r}{0.6\textwidth}
\centering
\vspace{-4mm}
\caption{Inference time and memory usage on DLC-Bench~\cite{lian2025describe} (Image) and HC-STVG~\cite{tang2021human} (Video). We report per-item inference time (s/item) and peak GPU memory (GB).}
\vspace{1mm}
\label{tab:time}
\resizebox{0.99\linewidth}{!}{
\begin{tabular}{l|cc|cc}
\toprule
& \multicolumn{2}{c}{\textbf{DLC-Bench}} & \multicolumn{2}{c}{\textbf{HC-STVG}} \\
\cmidrule(lr){2-3}\cmidrule(lr){4-5}
\textbf{Model} & \textbf{Infer time(s)$\downarrow$} & \textbf{Memory(GB)$\downarrow$} & \textbf{Infer time(s)$\downarrow$} & \textbf{Memory(GB)$\downarrow$} \\
\midrule
DAM-3B~\cite{lian2025describe} & 1.29 & 7.8  & 2.68 & 10.4 \\
PAM-3B~\cite{lin2025perceive} & 1.09 & 9.4 & 1.51 & 12.7  \\
\rowcolor{blue!5}\textbf{\ours-2B}  & 1.04 & 13.2 & 0.82 & 24.6 \\
\rowcolor{blue!5}\textbf{\ourslite-2B} & \textbf{0.88} & \textbf{4.9}  & \textbf{0.68} & \textbf{5.2}  \\
\rowcolor{blue!5}\textbf{\ours-7B}  & 1.44 & 25.1 & 1.25 & 36.9 \\
\rowcolor{blue!5}\textbf{\ourslite-7B} & 1.10 & 15.8 & 0.74 & 17.6 \\
\bottomrule
\end{tabular}}
\vspace{-2.5mm}
\end{wraptable}

The FLOPs and memory usage of our model are reported in Table~\ref{tab:flops}. In the video setting, we uniformly sample 20 frames for each video to ensure a fair comparison.  As shown, the Object-Only Framework significantly reduces computational and memory demands. For instance, 
with video inputs,
\ours-2B requires 11.15T FLOPs and 24.6GB of GPU memory, whereas the object-only variant \ours-2B-Lite reduces the cost to merely 0.11T FLOPs and 5.1GB of memory. 
Similar reductions are also observed for larger models and in the image input setting.
These results highlight that the Object-Only Framework is highly efficient in minimizing computational overhead and memory consumption, providing a scalable and cost-effective solution for large-scale applications without compromising performance.
As shown in Table~\ref{tab:time}, we further provide a detailed comparison of inference time and memory usage with DAM~\cite{lian2025describe} and PAM~\cite{lin2025perceive}. Notably, \ours-2B achieves significant reductions in both metrics, particularly in the video setting.

\begin{figure*}[t]
  \centering
\includegraphics[width=0.99\linewidth]{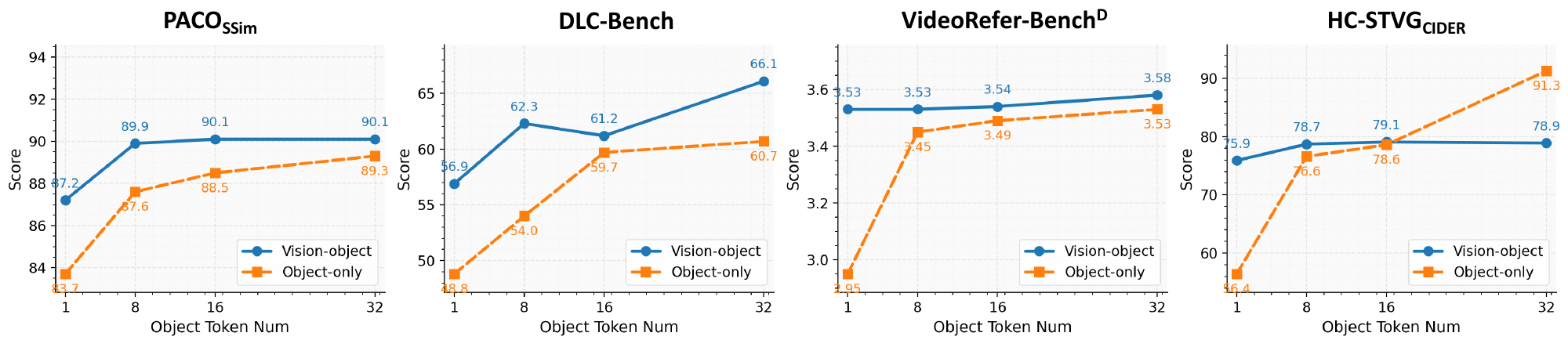}
\vspace{1.0mm}
   \caption{Effects of object token scaling across four typical benchmarks, including PACO, DLC-Bench, VideoRefer-Bench\textsuperscript{D}, and HC-STVG. We evaluate the impact of varying object token numbers (1, 8, 16, 32) under two model configurations: Vision-Object and Object-Only.}
   \label{fig:token-scaling}
   \vspace{-2mm}
\end{figure*}

\begin{table}[t]
\begin{minipage}{0.48\linewidth}
    \caption{Ablation results on different object token numbers with both Vision-Object Framework and Object-Only Framework.}
    \centering
    \resizebox{\columnwidth}{!}{
    \begin{tabular}{c|cccc}
        \toprule
     \textbf{Token Num.} & \textbf{PACO$_\text{SSim}$} & \textbf{DLC-Bench} & \textbf{VideoRefer$^\text{D}$} & \textbf{HC-STVG$_\text{CIDER}$}\\
     \midrule
     \rowcolor[HTML]{F2F2F2} \multicolumn{5}{l}{\textbf{Vision-Object Framework}} \\ \midrule
        1  & 87.2  & 56.9 & 3.53  & 75.9   \\
        8  & 89.9  & 62.3  & 3.53  &   78.7  \\
        16 & \textbf{90.1} & 61.2  & 3.54  &  \textbf{79.1}   \\ 
        \rowcolor{blue!5}32 & \textbf{90.1}  & \textbf{66.1}  & \textbf{3.58}  &  78.9  \\ 
        \midrule
        \rowcolor[HTML]{F2F2F2} \multicolumn{5}{l}{\textbf{Object-Only Framework}} \\ \midrule
        1  & 83.7  &  48.8 & 2.95  & 56.4   \\
        8  &  87.6 & 54.0  & 3.45  & 76.6   \\
        16 & 88.5  &  59.7 & 3.49  & 78.6    \\ 
        \rowcolor{blue!5}32 &  \textbf{89.3} & \textbf{60.7}  & \textbf{3.53}  & \textbf{91.3}   \\ 
        \bottomrule
    \end{tabular}}
    \label{tab:token_num}
\end{minipage}
\begin{minipage}{0.49\linewidth}
\centering\small
\caption{Ablation results for the design of the Scale-Adaptive Object Tokenizer across both image and video benchmarks.}
    \centering
    \resizebox{\columnwidth}{!}{
    \begin{tabular}{l|ccccc}
        \toprule
     \textbf{Method} & \textbf{LVIS$_\text{Avg}$} & \textbf{DLC-Bench} & \textbf{VideoRefer$^\text{D}$} & \textbf{HC-STVG$_\text{CIDER}$}\\
     \midrule
     Mask Pooling & 79.4 & 56.8 & 3.50 & 76.6\\
     w/o expansion & 81.0 & 65.4 & 3.56 & 78.0 \\
     w/o position emb. & 86.0 & 64.3 & 3.52 & 77.2 \\
     \rowcolor{blue!5}\textbf{Ours} & \textbf{86.2} & \textbf{66.1} & \textbf{3.58} & \textbf{78.9} \\
        \bottomrule
    \end{tabular}}
    \label{tab:ablation-tokenizer}
\caption{Ablation results on the design of Object-Centric Infusion Module.}
    \centering
    \resizebox{\columnwidth}{!}{
    \begin{tabular}{cc|cccc}
        \toprule
     \textbf{L-Attn} & \textbf{G-Attn} & \textbf{LVIS$_\text{SSim}$} & \textbf{DLC-Bench} & \textbf{VideoRefer$^\text{D}$} & \textbf{HC-STVG$_\text{CIDER}$}\\
     \midrule
     \ding{55} & \ding{55} & 85.0 & 59.0 & 3.37 & 69.6\\
     \ding{51} & \ding{55} & 85.2 & 60.5 & 3.46 & 73.7 \\
     \ding{55} & \ding{51} & 88.2 & 60.6 & 3.48 & 77.2 \\
     \rowcolor{blue!5}\ding{51} & \ding{51} & \textbf{89.4} & \textbf{60.7} & \textbf{3.53} & \textbf{91.3}\\
        \bottomrule
    \end{tabular}}
    \label{tab:ablation-fusion}
\end{minipage}
\end{table}

\subsection{Ablation Study}
\begin{wrapfigure}{r}{0.49\textwidth}
  \centering
\includegraphics[width=0.9999\linewidth]{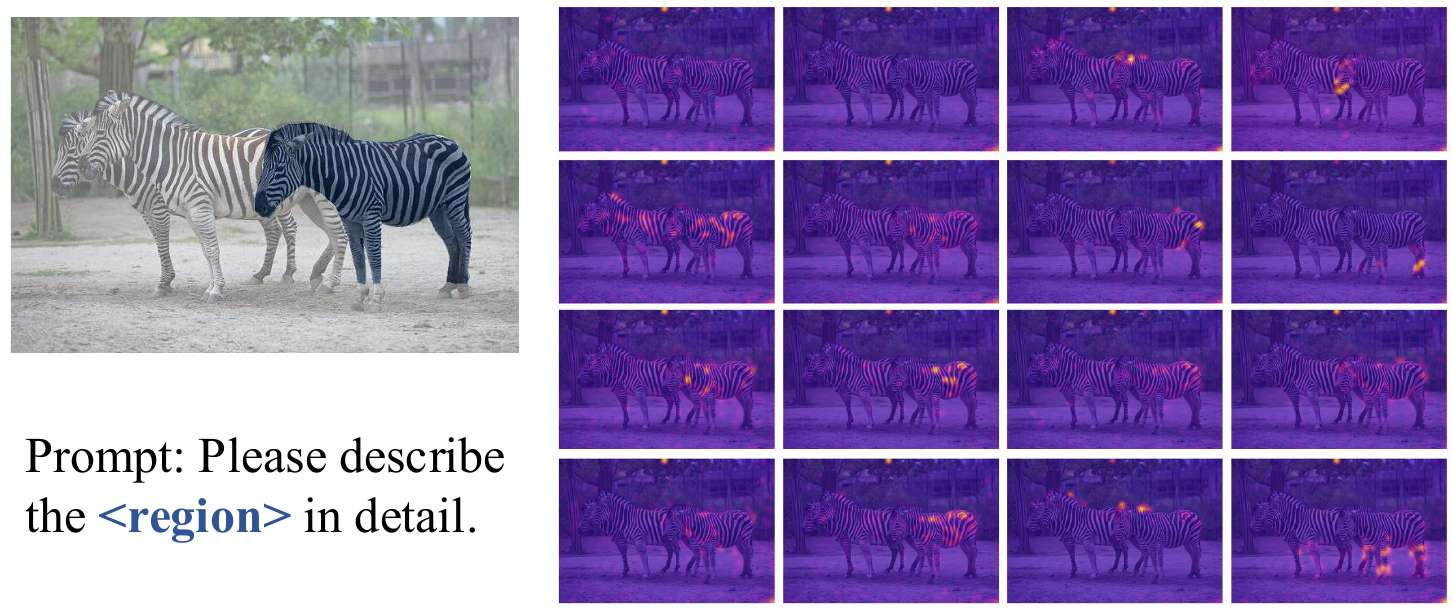}
   \caption{Visualization of attention map between object tokens and image tokens with 16 tokens.}
   \vspace{-3.5mm}
   \label{fig:attn_map}
\end{wrapfigure}
\textbf{Scaling Object Tokens.} We study how the number of object tokens in our scale-adaptive tokenizer influences both the \emph{Vision–Object} and \emph{Object-Only} frameworks. The results are presented in Table~\ref{tab:token_num} and Fig.~\ref{fig:token-scaling}. 
For the Vision–Object Framework, increasing the number of object tokens from 1 to 8 yields the most substantial gains across benchmarks. 
Beyond this point, improvements largely plateau: PACO\textsubscript{SSim} and VideoRefer-Bench\textsuperscript{D} show little change beyond 8 tokens, HC-STVG drops marginally, while DLC-Bench continues to benefit up to 32 tokens. In contrast, the Object-Only Framework exhibits consistent improvements as the number of tokens increases. 
Adding more tokens progressively narrows the performance gap relative to the Vision–Object model, and on HC-STVG, it even surpasses the latter when using 32 tokens. These findings highlight the complementary role of global vision tokens, which provide scene-level context and allow the Vision–Object model to achieve strong results with relatively few object tokens.
Conversely, the Object-Only model relies more heavily on a larger token budget to capture fine-grained object details and relational information.
We further explore the role of the number of object tokens by visualizing the attention patterns between different object tokens and image tokens in Fig.~\ref{fig:attn_map} (using 16 tokens as an example). The visualization reveals that different object tokens focus on distinct regions of the objects, thereby supplementing detailed information.

\begin{table*}[t]
    \caption{Ablation results on the impact of various training data types. 
    We utilize  SSim  for the LVIS benchmark and METEOR for the HC-STVG, and the average scores for the remaining benchmarks.}
    \vspace{1.0mm}
    \centering
    \resizebox{\columnwidth}{!}{
    \begin{tabular}{l|c|cc|ccc|cc}
        \toprule
        \multirow{2}{*}{\textbf{Data}} & \multirow{2}{*}{\textbf{\#Samples}} & \multicolumn{2}{c|}{\textbf{Image-Region-Bench}} & \multicolumn{3}{c|}{\textbf{Video-Region-Bench}} & \multicolumn{2}{c}{\textbf{General-Bench}} \\
        \cmidrule(lr){3-4}\cmidrule(lr){5-7}\cmidrule(lr){8-9}
        & & \textbf{LVIS} & \textbf{DLC-Bench} & \textbf{HC-STVG} & \textbf{VideoRefer$^\text{D}$} & \textbf{VideoRefer$^\text{Q}$}  & \textbf{MVBench} & \textbf{POPE} \\
     \midrule
     Region Recognition & 390K & 89.6 & 61.2 & 11.9 & 2.94 & 72.3  & 60.3& 87.3 \\
     + Image Detailed Cap. & 860K & 89.7 & 66.4 & 13.0 & 2.97 & 71.9  & 58.7 & 88.2 \\
     + Video Detailed Cap. & 180K & 89.7 & 66.0 & 19.1 & \textbf{3.69} & 74.8  & 61.9 & 88.0 \\
     + Region QA & 560K & 89.7 & \textbf{66.6} & \textbf{19.6} & 3.62 & 75.8  & 61.6 & 83.9 \\
     \rowcolor{blue!5} + General QA & 300K & \textbf{89.8} & 66.1 & 19.5 & 3.58 & \textbf{76.5}  & \textbf{63.4} & \textbf{88.7} \\
        \bottomrule
    \end{tabular}}
    \vspace{-1mm}
    \label{tab:ablation-data}
\end{table*}

\begin{wrapfigure}{r}{0.52\textwidth}
  \centering
  \vspace{-3mm}
\includegraphics[width=0.9999\linewidth]{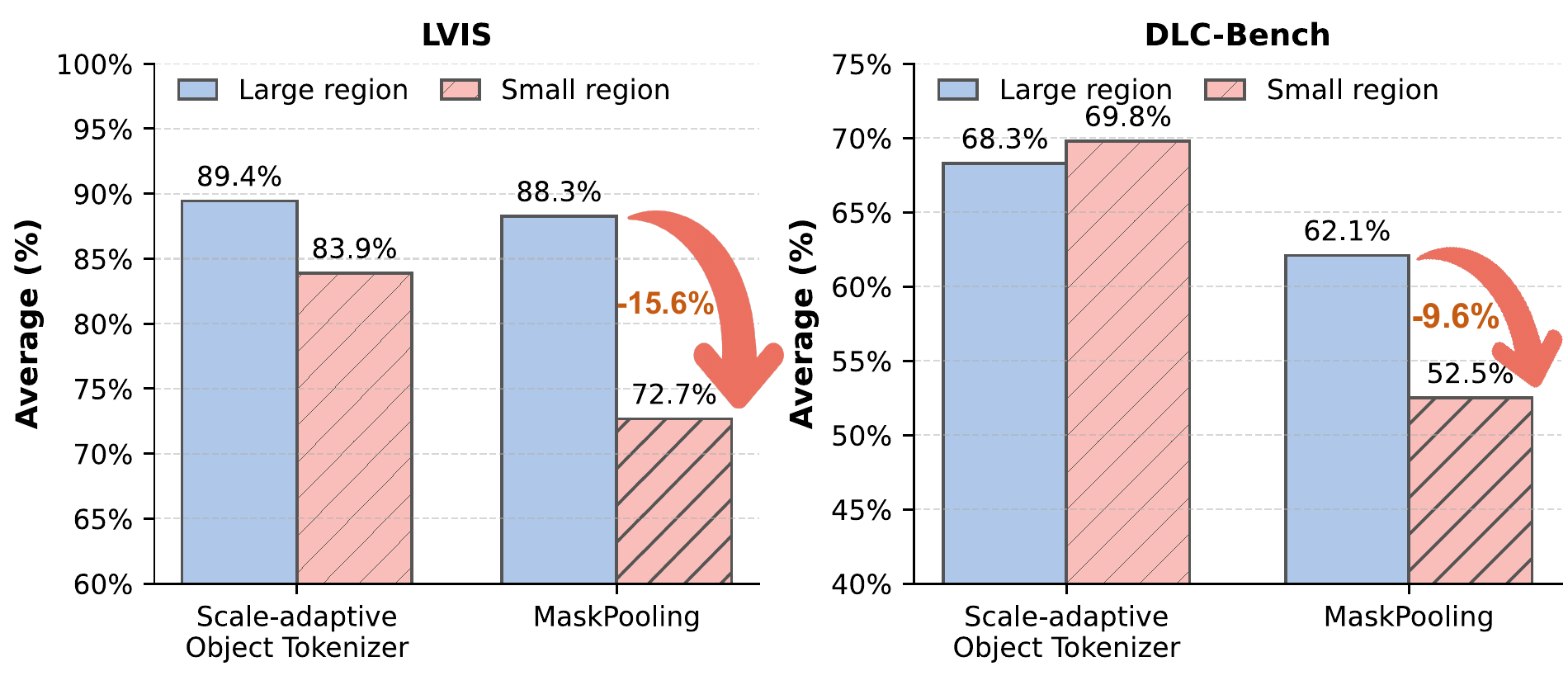}
\vspace{-2.0mm}
   \caption{Performance comparisons between the proposed Scale-Adaptive Object Tokenizer (SAOT) and MaskPooling on LVIS and DLC-Bench with large and small regions.  Our SAOT demonstrates consistently strong performance across both region sizes, while MaskPooling suffers significant degradation on small regions, highlighting the importance of scale-aware object representation.
   }
   \vspace{-3mm}
   \label{fig:performance_diff}
\end{wrapfigure}

\noindent\textbf{Design of Scale-Adaptive Object Tokenizer (SAOT).} We conduct an in-depth analysis of the design in the Scale-adaptive Object Tokenizer, as illustrated in Table~\ref{tab:ablation-tokenizer}. 
First, we compare our design with the vanilla Mask Pooling method~\cite{yuan2024osprey,yuan2025videorefer}. Our tokenizer achieves significant gains on both image and video benchmarks, outperforming the baseline with 6.8\% on LVIS, 8.6\% on DLC-Bench and 1.4\% on HC-STVG. 
To further investigate its efficacy, we divide the regions into two groups based on pixel count: small regions (\textless2000 pixels) and large regions (\(\geq\)2000 pixels).
As depicted in Fig.~\ref{fig:performance_diff}, the performance gap is particularly pronounced for smaller regions,
with improvements of 15.6\% on  LVIS and 9.6\% on DLC-Bench. These results clearly showcase the effectiveness of our design in preserving object details, especially in scenarios involving tiny objects.

We further analyze the impact of the expansion operation, which incorporates surrounding context after region cropping to enrich feature representations.
As shown in Table~\ref{tab:ablation-tokenizer}, omitting this expansion results in  a noticeable decline in performance, particularly on image benchmarks, with decreases of 5.2\% on LVIS and 0.7\% on DLC-Bench. These results underscore the key role of contextual information in enhancing region-level feature extraction.

Finally, we examine the design of position embedding, which incorporates relative positional features into object tokens. As shown in Table~\ref{tab:ablation-tokenizer}, this design yields improvements in both image and video benchmarks, particularly for tasks requiring detailed descriptions. 
These tasks necessitate not only accurate category recognition but also a coherent understanding of each object’s spatial location within the image or video sequence.

\noindent\textbf{Design of Object-Centric Infusion (OCI) Module.} 
We analyze the effects of \textit{Local-to-Object Attention} (L-Attn) and \textit{Global-to-Object Attention} (G-Attn) within the Object-Centric Infusion Module. Table~\ref{tab:ablation-fusion} presents the results.
The baseline uses only object features, without either L-Attn or G-Attn, meaning the model can only ``see'' the object itself, without contextual cues. Introducing local context through L-Attn yields consistent improvements across all benchmarks, confirming that nearby contextual information aids in disambiguating object understanding. Adding global context via G-Attn leads to even larger gains, highlighting the importance of scene-level cues when interpreting small or ambiguous regions. 
When both mechanisms are combined, performance reaches its highest level: +4.4\% on LVIS, +1.7\% on DLC-Bench, +0.16 on VideoRefer-D, and a striking +21.7\% on HC-STVG. These results confirm that local and global contexts are complementary, where local cues refine details, while global cues provide holistic scene information, together enabling more effective object-centric representation.

\noindent\textbf{Impact of Diverse Training Data.} To evaluate the effectiveness of the datasets collected in our \ours-2.2M, we classify the datasets we used into six types: Region Recognition, Image Detailed Caption, Video Caption, Region QA and General QA. Table~\ref{tab:ablation-data} reports results across diverse benchmarks, spanning region- and scene-level, image- and video-level, QA- and description-level tasks. 
Starting with only the region recognition datasets, the model exhibits basic category cognition with 89.6\% on LVIS, which is relatively easy, but struggles on tasks requiring detailed descriptions or QA.
Incorporating image and video captioning data substantially enhances captioning performance while preserving region recognition ability.
The inclusion of Region QA data further enhances QA performance, most notably on VideoRefer-Bench$^Q$.
Lastly, incorporating General QA data strengthens the model's QA capabilities without impairing other tasks,
thereby mitigating the risk of catastrophic forgetting.

\section{Conclusion}
We presented~\ours, a unified region-level MLLM framework designed to support fine-grained spatio-temporal object-centric understanding across images and videos with arbitrary granularity. 
By introducing the Scale-Adaptive Object Tokenizer (SAOT), \ours~generated compact and semantically rich object representations from free-form regions. Building upon empirical analysis of attention patterns within LLMs, we further developed \ours-Lite, an efficient Object-Only Framework that employs an Object-Centric Infusion module to pre-fuse global context into object tokens, significantly improving efficiency without sacrificing accuracy. 
To support robust training, we curated \ours-2.2M, a high-quality object-centric instruction dataset. 
Extensive experiments across diverse tasks, ranging from captioning and recognition to complex reasoning, demonstrated \ours's state-of-the-art performance  with fewer training samples.
Meanwhile, the \ours-Lite~variant offers comparable accuracy with notable efficiency gains, highlighting the practicality and scalability of our proposed framework.

\bibliographystyle{assets/plainnat}
\bibliography{paper}


\end{document}